\documentclass{article}
\PassOptionsToPackage{numbers, sort&compress}{natbib}

\usepackage[preprint]{neurips_2025}

\usepackage{natbib}
\usepackage[T1]{fontenc}    
\usepackage{hyperref}       
\usepackage{url}            
\usepackage{booktabs}       
\usepackage{amsfonts}       
\usepackage{nicefrac}       
\usepackage{microtype}      
\usepackage{xcolor}         
\usepackage{graphicx}
\usepackage{multirow}
\usepackage{multicol}
\usepackage{booktabs}
\usepackage{amsmath}
\usepackage{subfigure}
\usepackage{todonotes}
\usepackage{tcolorbox}
\usepackage{subcaption}
\usepackage{CJKutf8}    
\usepackage{bbm}
\usepackage{pifont}

\usepackage[utf8]{inputenc} 
\usepackage[whole]{bxcjkjatype}
\newcommand{\Chinese}[1]{{\begin{CJK*}{UTF8}{gbsn}#1\end{CJK*}}}

\title{Do LLMs Need to Think in One Language? \\ Correlation between Latent Language \\ and Task Performance}

\author{%
  Shintaro Ozaki${}^{\alpha, \beta}$ \hspace{3pt} 
  Tatsuya Hiraoka${}^{\gamma}$ \hspace{3pt} 
  Hiroto Otake${}^{\alpha, \beta}$ \hspace{3pt}  \\[5pt]
  \textbf{Hirki Ouchi}${}^{\alpha, \eta}$ \hspace{3pt}
  \textbf{Masaru Isonuma}${}^{\beta, \delta, \epsilon}$ \hspace{3pt}
  \textbf{Benjamin Heinzerling}${}^{\eta, \delta}$ \hspace{3pt}
  \textbf{Kentaro Inui}${}^{\gamma, \delta, \eta}$ \\[5pt]
  \textbf{Taro Watanabe${}^{\alpha}$} \hspace{3pt}
  \textbf{Yusuke Miyao${}^{\epsilon, \beta}$} \hspace{3pt}
  \textbf{Yohei Oseki${}^{\epsilon}$} \hspace{3pt}
  \textbf{Yu Takagi${}^{\theta}$} \\[10pt]
  ${}^{\alpha}$NAIST \hspace{5pt}
  ${}^{\beta}$NII LLMC \hspace{5pt}
  ${}^{\gamma}$MBZUAI \hspace{5pt}
  ${}^{\eta}$RIKEN \\[5pt]
  ${}^{\delta}$Tohoku University \hspace{5pt}
  ${}^{\epsilon}$The University of Tokyo \hspace{5pt}
  ${}^{\theta}$Nagoya Institute of Technology \\[10pt]
\texttt{ozaki.shintaro.ou6@naist.ac.jp　　takagi.yu@nitech.ac.jp}
}

\begin{document}
\maketitle

\begin{abstract}
Large Language Models (LLMs) are known to process information using a proficient internal language consistently, referred to as latent language, which may differ from the input or output languages.
However, how the discrepancy between the latent language and the input and output language affects downstream task performance remains largely unexplored.
While many studies research the latent language of LLMs, few address its importance in influencing task performance.
In our study, we hypothesize that thinking in latent language consistently enhances downstream task performance.
To validate this, our work varies the input prompt languages across multiple downstream tasks and analyzes the correlation between consistency in latent language and task performance.
We create datasets consisting of questions from diverse domains such as translation and geo-culture, which are influenced by the choice of latent language.
Experimental results across multiple LLMs on translation and geo-culture tasks, which are sensitive to the choice of language, indicate that maintaining consistency in latent language is not always necessary for optimal downstream task performance.
This is because these models adapt their internal representations near the final layers to match the target language, reducing the impact of consistency on overall performance.
\end{abstract}

\section{Introduction}
\label{introduction}
Large Language Models (LLMs) are known to process information using a proficient internal language consistently, referred to as latent language~\cite{wendler-etal-2024-llamas, zhong2024beyond}, which may differ from the input or output languages.
For instance, Llama2~\cite{touvron2023llama}, which is primarily trained on English-centric corpora, employs English as a latent language during inference, shaping its internal representations accordingly~\cite{wendler-etal-2024-llamas}.
Similarly, models like LLM-jp~\cite{aizawa2024llm}, which are trained on both Japanese and English data, tend to think in Japanese during their internal reasoning processes~\cite{zhong2024beyond}.
This pattern suggests that LLMs tend to adopt their proficient latent language shaped by their training data distribution.

\begin{figure}
    \centering
    \includegraphics[width=0.98\linewidth]{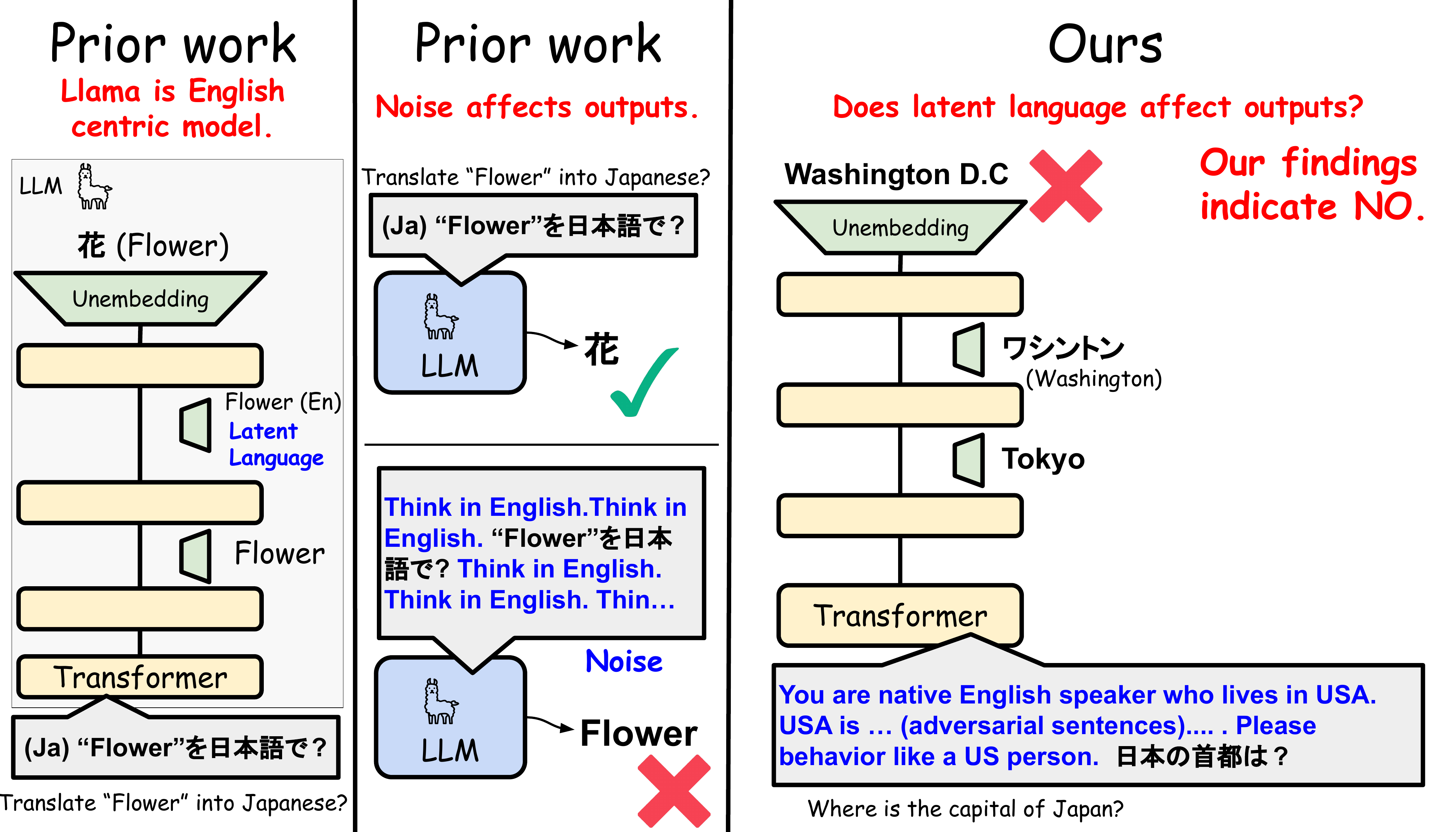}
    \caption{An overview of our research and differences from prior works.}
    \label{fig:top}
\end{figure}

However, the correlation between consistency in latent language and downstream task performance remains unexplored.
In particular, it is unclear how inputs in a less familiar language affect internal reasoning and, consequently, the accuracy of outputs as shown in Figure~\ref{fig:top}.
This perspective is critical for understanding the linguistic priors and stability of reasoning in multilingual LLMs~\cite{yong2025crosslingualreasoningtesttimescaling}.

In our study, we hypothesize that disrupting thought in latent language has an influence on downstream task performance.
To evaluate this, we deliberately disrupt the internal linguistic consistency of models by injecting adversarial prompts composed of multiple languages into the inputs and quantitatively evaluate how these changes affect downstream task performance.
To understand this correlation, we evaluate several LLMs known to possess their proficient latent language, systematically varying both the types and proportions of adversarial prompts' languages.
Additionally, we introduce a formal evaluation metric which we define as Latent Language Consistency Score (LLC Score) for quantifying the consistency in latent language within models.

Regarding tasks, there are no existing datasets that quantitatively evaluate our research focus.
Thus, our study constructs diverse question datasets inspired by prior research~\cite{wendler-etal-2024-llamas, zhong2024beyond, sakai-etal-2024-mcsqa, ozaki2024bqa}, specifically targeting domains where consistency in latent language is greatly affected by their accuracy.
In our work, we specifically focus on translation and geo-culture tasks, allowing for precise evaluation.

Experimental results on our created datasets, conducted across multiple LLMs trained in different languages, reveal that models often rely on a consistent latent language during inference.
However, even when they do so, they do not always achieve optimal performance, contradicting our initial hypothesis in both tasks.
In the translation task, the inclusion of adversarial prompts in less familiar languages from their proficient latent language degrades accuracy.
This suggests that linguistic noise in the inputs disrupts the internal reasoning process and destabilizes the model's internal structures.

\section{Related Work}
\label{related-work}
\subsection{Latent Language of LLMs}
\label{latent-language-of-llms}
LLMs are known to exhibit behavior strongly influenced by the languages prevalent in their pre-training data.
For example, Wendler et al~\cite{wendler-etal-2024-llamas} demonstrate that Llama2~\cite{touvron2023llama}, which is primarily trained on English data, tends to rely on English for internal reasoning as a latent language, while Zhong et al~\cite{zhong2024beyond} report that LLM-jp models~\cite{aizawa2024llm} trained on both Japanese and English exhibit internal representations toward Japanese language.
These findings suggest that LLMs often develop a proficient latent language~\cite{schut2025multilingual} for internal processing, reflecting inherent linguistic representations~\cite{haemmerl-etal-2023-speaking, gallegos2023bias} shaped by their data.

Additionally, the choice of prompt language has been shown to strongly affect model outputs and performance in multilingual settings.
For instance, Huang et al~\cite{huang-etal-2023-languages} found that prompts in different languages lead to substantial variations in generated outputs, i.e., accuracy.
Other studies~\cite{shi2022language, lai-nissim-2024-mcot} have analyzed the impact of multilingual Chain-of-Thought~\cite{wei2022chain} prompting on reasoning processes and performance.

However, these investigations primarily focus on surface level changes in outputs, such as response accuracy~\cite{xue-etal-2021-mt5}, without directly validating the latent language or how shifts in internal reasoning languages might affect downstream task performance.

\subsection{Robustness of LLMs to Adversarial Prompts}
Prior work has explored the robustness of models to external linguistic noise, i.e., adversarial prompts. 
For example, Goel et al~\cite{goel-etal-2021-robustness} examined the impact of various syntactic and lexical affections, while Khashabi et al~\cite{khashabi-etal-2020-bang} demonstrated that inserting semantically unrelated sentences can significantly reduce the output reliability of question answering models as shown in the center of Figure~\ref{fig:top}.
However, these studies primarily focus on how noise disrupts outputs, while leaving unexplored the potential shifts in the latent languages that models rely on during reasoning.

In contrast to those contributions, our study aims to systematically analyze how the latent language of LLMs influences output robustness and downstream task performance, addressing a critical gap in our understanding of language dependent reasoning.
By exploring the correlation between internal linguistic shifts and downstream task performance, we provide a comprehensive framework for evaluating the stability of internal reasoning in multilingual LLMs.

\subsection{Visualization of Internal Thought in LLMs}
\label{visualize-internal-thought-of-llms}
LogitLens~\cite{nostalgebraist2020logitlens, tunedlens} enables the calculation of logits, probabilities, and corresponding tokens at each intermediate layer of Transformer~\cite{vaswani2017attention} decoder models by projecting the hidden states through the unembedding matrix.
This approach makes it possible to analyze what the model is ``thinking'' and which language it is using at each layer of the Transformer~\cite{wendler-etal-2024-llamas}.
In our study, we utilize this approach to visualize the internal reasoning of the model, describing more details in Appendix~\ref{app:mathmetical-approach-for-logitlens}.
By extracting logits through the application of the unembedding matrix to the outputs of each layer, we can visualize tokens, compute probabilities, and further calculate the KL divergence between adjacent layers to define metrics for language consistency and robustness.

\section{Analysis Method}
\label{analysis-method}
In our study, we analyze how consistently thinking in a latent language affects downstream task performance, which we refer to as robustness.
To analyze this correlation, we first identify the latent language used in the model's intermediate layers during inference.
We then define specific metrics for both consistency in latent language and robustness, and evaluate whether reasoning in the latent language leads to optimal performance.
As we described in Section~\ref{introduction}, our hypothesis is disrupting the latent language should lead to a corresponding drop in robustness.

\subsection{Latent Language Detection}
\label{latent-language-detection}
As described in Section~\ref{latent-language-of-llms}, we call the language used internally by the model for reasoning as the latent language.
Inspired by prior work~\cite{wendler-etal-2024-llamas, zhong2024beyond, schut2025multilingual}, we use LogitLens~\cite{nostalgebraist2020logitlens}, introduced in Section~\ref{visualize-internal-thought-of-llms}, to extract tokens from the logits obtained at intermediate layers.
We then apply a language identification library\footnote{\url{https://github.com/saffsd/langid.py}.} to classify the extracted tokens as Japanese, Chinese, or English.


\subsection{Latent Language Consistency Score (LLC Score)}
\label{pivot-language-consistency}
We define the LLC Score to quantify how consistently a model relies on a latent language throughout its intermediate layers.
Our hypothesis is that if the model frequently shifts its internal reasoning across different latent languages, such behavior indicates a lack of coherence in internal processing.  
In contrast, when the same latent language is consistently used across layers, information flows more smoothly through the model, leading to higher consistency.

To capture this behavior, we define a score, $\text{Score}(v; \theta)$, for each candidate language $v \in \mathcal{V}$, which evaluates how stably the model internally uses that language.  
This score depends on two key components:
The change in internal representations between adjacent layers (KL divergence);
The output probability that the model uses language $v$ at each layer. (Applied by LogitLens~\cite{nostalgebraist2020logitlens}.)

When the model prefers a latent language $v$ at layer $l$ but switches to the different dominant language at layer $l+1$, i.e., $v_{l+1}^{*(\theta)} \neq v$, such a transition suggests a disruption in consistency for $v$.  
The function $\text{Score}(v; \theta)$ averages the degree of such disruptions:

\begin{equation}
\text{Score}(v; \theta) = \frac{\displaystyle \sum_{l=1}^{L-1} \left( P_{l,v}^{(\theta)} \cdot KL_{l,l+1}^{(\theta)} + KL_{l,l+1}^{(\theta)} \cdot P_{l+1,v}^{(\theta)} \right) \cdot \mathbbm{1}(v_{l+1}^{*(\theta)} \neq v)}{\displaystyle \sum_{l=1}^{L-1} \left( P_{l,v}^{(\theta)} + P_{l+1,v}^{(\theta)} \right) \cdot \mathbbm{1}(v_{l+1}^{*(\theta)} \neq v)}
\end{equation}

We define $P_{l,v}^{(\theta)}$ as the probability that the model uses latent language $v$ at layer $l$.
The term $KL_{l,l+1}^{(\theta)}$ denotes the KL divergence between the output distributions at layers $l$ and $l+1$.
We denote the most likely language at layer $l$ as $v_{l}^{*(\theta)} = \arg\max_{u \in \mathcal{V}} P_{l,u}^{(\theta)}$, where $u$ represents a candidate language. 
The indicator function $\mathbbm{1}(v_{l+1}^{*(\theta)} \neq v)$ returns 1 if the most likely language at layer $l+1$ differs from $v$.

With LogitLens~\cite{nostalgebraist2020logitlens} as we described in Section~\ref{visualize-internal-thought-of-llms}, $KL_{l,l+1}^{(\theta)}$ enables extraction of logits from each intermediate layer, and $v_{l}^{*(\theta)} = \arg\max_{u \in \mathcal{V}} P_{l,u}^{(\theta)}$ identifies the language of the token with the highest output probability determined by Section~\ref{latent-language-detection}.

The score aggregates internal disruptions related to language $v$, weighted by the model's output probability in using that language.
Then, the overall LLC Score for model $\theta$ is defined as the minimum of the consistency scores across all candidate languages:

\begin{equation}
\text{LLC Score}(\theta) = \min_{v \in \mathcal{V}} \text{Score}(v; \theta)
\end{equation}

A lower LLC Score suggests that the model maintains stable internal representations using at least one latent language, just as lower KL divergence indicates better layerwise consistency.

Following prior work~\cite{nostalgebraist2020logitlens, belrose2023eliciting}, we restrict our analysis to the latter half of the model's intermediate layers, where semantic reasoning tends to consolidate.

\subsection{Robustness against Adversarial Prompts}
\label{robustness-against-adversarial-promps}
We define accuracy, i.e., the model's outputs, as a measure of robustness.
Our hypothesis, as described in Section~\ref{analysis-method}, is that when the latent language of the model is disrupted, the robustness in downstream tasks should also be similarly affected.

\subsection{Adversarial Prompts}
\label{adversarial-prompts}
To guide the latent language processing of models, we introduce adversarial prompts.
Drawing on prior work~\cite{khashabi-etal-2020-bang, goel-etal-2021-robustness} that highlights the accuracy degradation caused by adversarial inputs, we aim to disrupt consistency in latent language as well.
Specifically, we create prompts that describe the culture, background, and history of Japan, English, Chinese-speaking regions, and are followed by an instruction like, ``Based on the above, act as if you are in Chinese.'' as shown in Appendix~\ref{app:detailed-prompts}.
This approach tests whether inducing models to think in the prompt language can disrupt consistency in latent language and affect robustness.

\section{Dataset Creation}
\label{dataset-creation}
We choose cloze style tasks~\cite{wendler-etal-2024-llamas, zhong2024beyond, petroni-etal-2019-language}, where the target is to predict a specific word or phrase that fills a blank space ('\_') in a sentence~\cite{taylor1953cloze}.
This format is particularly suitable for LogitLens analysis~\cite{nostalgebraist2020logitlens}, as it focuses on specific token predictions, allowing us to more precisely interpret intermediate representations.
In prior research, no cloze task datasets specifically designed to evaluate the impact of linguistic consistency on model robustness have been developed.
To address this issue, we construct a dataset, following prior studies~\cite{sakai-etal-2024-mcsqa, ozaki2024bqa} that demonstrate the feasibility of semi-automatically generating QA pairs using LLMs.
Our work focuses on translation and geo-culture domains, which are particularly sensitive to the choice of language.
We split into two distinct steps, following the method described in previous work~\cite{sakai-etal-2024-mcsqa, ozaki2024bqa}, which emphasizes the importance of separating question generation from model inference.
This approach ensures that the generated questions are not biased by the models being evaluated, providing a cleaner test.

\subsection{STEP 1: Generate Questions}
\label{step1-generate-questions}
Our work uses prompts like ``Generate cloze task questions and their answers for translation.'' to generate cloze task questions~\cite{taylor1953cloze}.
To ensure diversity and avoid overfitting to a single prompt structure~\cite{samuel2024personagym}, we prepare a list of several dozen category labels as shown in Appendix~\ref{app:questions-diversity-in-datasets}.
When generating the questions, these categories are selected at random, encouraging the model to produce a wide range of questions and reducing the risk of repetitive patterns.
We also calculate Self-BLEU~\cite{zhu2018texygen} as a diversity metric to quantitatively evaluate; the results of which are presented in Table~\ref{tab:self-bleu} in the Appendix~\ref{app:questions-diversity-in-datasets}.
These results confirm that our dataset exhibits diverse questions, ensuring its reliability.
For question generation, we use GPT-4o~\cite{hurst2024gpt}, the highest performing model.
Appendix~\ref{app:detailed-prompts} provides the specific prompts used for this process.

\subsection{STEP 2: Filter Questions}
\label{step2-filter-questions}
After generating the initial set of cloze task questions in STEP 1, we perform an additional filtering step to ensure the quality and consistency of the dataset.
Specifically, we instruct GPT-4o to answer the generated questions and then compare the predicted answers with the original answers created during question generation.
Only the questions for which the model's predicted answer exactly matches the original gold answer are retained in the final dataset.

Additionally, we remove any questions that do not meet the following criteria.
(1) Single-Token Answers: The answer must be a single token, following the approach used in prior studies~\cite{petroni-etal-2019-language};
(2) Cloze Task Format: The question must adhere to the cloze task format, as defined in previous work~\cite{sakai-etal-2024-mcsqa}, meaning the prompt must clearly indicate a single missing word or phrase.

After this filtering process, we obtain a final dataset of 2,000 questions for each task.
This strict filtering ensures that the resulting dataset effectively captures whether consistency in latent language is disrupted.
Examples from the final dataset include translation tasks like ``Please translate flower into Japanese.\_, answer: '' and geo-culture tasks like ``Who is the prime minister of Japan?\_, answer: '' 
Appendix~\ref{app:examples-of-questions} provides further details on the dataset construction and filtering criteria.

\begin{table}[t]
    \centering
    \small
    \caption{All of settings we used in our experiments.}
    \begin{tabular}{cc}
        \toprule
       \textbf{Option} & \textbf{Candidates} \\
        \midrule
         Task & Translation, Geo-culture \\
         Language & English, Japanese, Chinese \\
         Data size & 2,000 for each\\
         Model & Qwen2.5, Gemma3, LLM-jp-3 \\
         Translation Task Language& \{En-Ja\}, \{En-Zh\}, \{Zh-En\}, \{Ja-En\} \\
         Geo-culture Task Language & \{En\}, \{Ja\}, \{Zh\} \\
         Adversarial Prompts Language  &  \{En\}, \{Ja\}, \{Zh\} \\
         Adversarial Prompts Ratio & \{None, 0.2, 0.4, 0.6, 0.8, 1.0\} \\
         \bottomrule
    \end{tabular}
    \label{tab:experimental-sesttins}
\end{table}

\section{Experiments}
The goal of our experiments is to examine how the consistency of latent language affects downstream task performance across multilingual LLMs.
We specifically aim to test the hypothesis that maintaining internal consistency in latent language contributes to higher task accuracy, especially when models encounter prompts containing unfamiliar languages.
Ideally, models that consistently think in their proficient latent language should maintain good robustness.

\subsection{Models}
\label{models}
For our experiments, we select models that are primarily pre-trained on English, Chinese, and Japanese data, including Gemma3~\cite{team2025gemma}, Qwen2.5~\cite{yang2024qwen2}, and LLM-jp~\cite{aizawa2024llm} respectively.
We select them to evaluate the impact of performing downstream tasks in languages that differ from each model's dominant latent language, providing a more comprehensive understanding of consistency and robustness.
Appendix~\ref{app:detailed-model-settings} provides detailed settings and model configurations.

\begin{table}[t]
\caption{
Correlation ($r$) between LLC Score and robustness in each ratio by geo-culture tasks.
}
    \small
    \centering
    \resizebox{\textwidth}{!}{
    \begin{tabular}{cccc ccccc ccccc}
    \toprule
\multirow{2.25}{*}{\textbf{Model}} & \multirow{2.25}{*}{\textbf{Question}} & \multirow{2.25}{*}{\textbf{Adversarial}} & \multicolumn{5}{c}{\textbf{LLC Score ($\downarrow$)}} & \multicolumn{5}{c}{\textbf{Robustness ($\uparrow$)}} & \multirow{2}{*}{\textbf{\(r\)}} \\
\cmidrule(lr){4-8}\cmidrule(lr){9-13}
 & & & \textbf{0.2} & \textbf{0.4} & \textbf{0.6} & \textbf{0.8} & \textbf{1.0} & \textbf{0.2} & \textbf{0.4} & \textbf{0.6} & \textbf{0.8} & \textbf{1.0} \\
\midrule
\multirow{8}{*}{LLM-jp-3} & \multirow{2}{*}{Ja} & Ja & $\text{Ja}_{0.06}$ & $\text{En}_{0.08}$ & $\text{Ja}_{0.09}$ & $\text{Ja}_{0.10}$ & $\text{Ja}_{0.11}$ & 0.27 & 0.26 & 0.27 & 0.24 & 0.24 & -0.82 \\
 & & En & $\text{Ja}_{0.09}$ & $\text{Ja}_{0.07}$ & $\text{Ja}_{0.11}$ & $\text{Ja}_{0.10}$ & $\text{Ja}_{0.11}$ & 0.13 & 0.22 & 0.13 & 0.06 & 0.04 & -0.83 \\
\cmidrule(lr){3-14}
 & \multirow{3}{*}{En} & Ja & $\text{Ja}_{0.06}$ & $\text{Ja}_{0.07}$ & $\text{En}_{0.12}$ & $\text{En}
_{0.12}$ & $\text{En}_{0.13}$ & 0.15 & 0.11 & 0.10 & 0.15 & 0.13 & -0.17 \\
 & & En & $\text{En}_{0.10}$ & $\text{En}_{0.11}$ & $\text{En}_{0.11}$ & $\text{En}_{0.12}$ & $\text{En}_{0.13}$ & 0.22 & 0.20 & 0.20 & 0.18 & 0.17 & -0.97 \\
 & & Zh & $\text{Ja}_{0.04}$ & $\text{En}_{0.12}$ & $\text{En}_{0.13}$ & $\text{En}_{0.13}$ & $\text{En}_{0.13}$ & 0.10 & 0.10 & 0.09 & 0.10 & 0.12 & -0.05 \\
\cmidrule(lr){3-14}
 & \multirow{2}{*}{Zh} & En & $\text{En}_{0.11}$ & $\text{En}_{0.12}$ & $\text{En}_{0.12}$ & $\text{Zh}_{0.09}$ & $\text{En}_{0.13}$ & 0.04 & 0.05 & 0.04 & 0.03 & 0.02 & 0.21 \\
 & & Zh & $\text{Ja}_{0.05}$ & $\text{Ja}_{0.06}$ & $\text{Ja}_{0.10}$ & $\text{En}_{0.11}$ & $\text{En}_{0.10}$ & 0.07 & 0.07 & 0.06 & 0.05 & 0.02 & -0.61 \\
\cmidrule(lr){2-14}
\multirow{8}{*}{Qwen2.5} & \multirow{2}{*}{Ja} & Ja & $\text{En}_{0.10}$ & $\text{En}_{0.09}$ & $\text{En}_{0.10}$ & $\text{En}_{0.09}$ & $\text{En}_{0.10}$ & 0.00 & 0.00 & 0.00 & 0.00 & 0.00 & N.A. \\
& & En & $\text{En}_{0.09}$ & $\text{En}_{0.09}$ & $\text{En}_{0.09}$ & $\text{En}_{0.10}$ & $\text{En}_{0.10}$ & 0.00 & 0.00 & 0.00 & 0.00 & 0.00 & N.A. \\
\cmidrule(lr){3-14}
& \multirow{3}{*}{En} & Ja & $\text{En}_{0.09}$ & $\text{En}_{0.09}$ & $\text{En}_{0.09}$ & $\text{En}_{0.09}$ & $\text{En}_{0.09}$ & 0.31 & 0.30 & 0.27 & 0.25 & 0.27 & -0.94 \\
& & En & $\text{En}_{0.09}$ & $\text{En}_{0.09}$ & $\text{En}_{0.09}$ & $\text{En}_{0.10}$ & $\text{En}_{0.10}$ & 0.33 & 0.27 & 0.25 & 0.29 & 0.32 & -0.25 \\
& & Zh & $\text{En}_{0.09}$ & $\text{En}_{0.09}$ & $\text{En}_{0.09}$ & $\text{En}_{0.10}$ & $\text{En}_{0.09}$ & 0.32 & 0.32 & 0.31 & 0.25 & 0.28 & -0.87 \\
\cmidrule(lr){3-14}
& \multirow{2}{*}{Zh} & En & $\text{En}_{0.09}$ & $\text{En}_{0.08}$ & $\text{En}_{0.08}$ & $\text{En}_{0.08}$ & $\text{En}_{0.09}$ & 0.00 & 0.00 & 0.00 & 0.00 & 0.00 & 0.21 \\
& & Zh & $\text{En}_{0.10}$ & $\text{En}_{0.10}$ & $\text{Zh}_{0.09}$ & $\text{Zh}_{0.09}$ & $\text{En}_{0.10}$ & 0.01 & 0.01 & 0.00 & 0.00 & 0.01 & 0.83 \\
\cmidrule(lr){2-14}
\multirow{8}{*}{Gemma3} & \multirow{2}{*}{Ja} & Ja & $\text{Zh}_{0.03}$ & $\text{Zh}_{0.01}$ & $\text{Zh}_{0.03}$ & $\text{En}_{0.03}$ & $\text{Zh}_{0.02}$ & 0.00 & 0.01 & 0.01 & 0.01 & 0.02 & -0.29 \\
& & En & $\text{En}_{0.02}$ & $\text{En}_{0.02}$ & $\text{Zh}_{0.02}$ & $\text{Zh}_{0.02}$ & $\text{En}_{0.02}$ & 0.00 & 0.01 & 0.01 & 0.00 & 0.00 & 0.90 \\
\cmidrule(lr){3-14}
& \multirow{3}{*}{En} & Ja & $\text{En}_{0.02}$ & $\text{En}_{0.03}$ & $\text{En}_{0.03}$ & $\text{En}_{0.02}$ & $\text{En}_{0.02}$ & 0.31 & 0.29 & 0.25 & 0.19 & 0.19 & 0.85 \\
& & En & $\text{En}_{0.02}$ & $\text{En}_{0.02}$ & $\text{En}_{0.02}$ & $\text{En}_{0.02}$ & $\text{En}_{0.02}$ & 0.29 & 0.27 & 0.17 & 0.14 & 0.12 & 0.98 \\
& & Zh & $\text{En}_{0.03}$ & $\text{En}_{0.03}$ & $\text{En}_{0.02}$ & $\text{En}_{0.02}$ & $\text{En}_{0.02}$ & 0.31 & 0.31 & 0.25 & 0.23 & 0.23 & 0.98 \\
\cmidrule(lr){3-14}
& \multirow{2}{*}{Zh} & En & $\text{En}_{0.02}$ & $\text{En}_{0.02}$ & $\text{En}_{0.02}$ & $\text{En}_{0.02}$ & $\text{En}_{0.02}$ & 0.00 & 0.01 & 0.00 & 0.00 & 0.00 & 0.83 \\
& & Zh & $\text{Zh}_{0.02}$ & $\text{Zh}_{0.02}$ & $\text{Zh}_{0.02}$ & $\text{Zh}_{0.02}$ & $\text{Zh}_{0.02}$ & 0.01 & 0.01 & 0.00 & 0.01 & 0.00 & -0.15 \\
    \bottomrule
    \end{tabular}
    } 
    \label{tab:result-table-geo}
\end{table}

\subsection{Datasets}
\label{datasets}
We utilize the dataset described in Section~\ref{dataset-creation}.
For translation tasks, possible Source-Target language pairs include \{En, Ja, Zh\}-\{En, Ja, Zh\}, but we exclude tasks where the source and target languages are the same like \{En\}-\{En\} or where the token language cannot be clearly identified, such as \{Ja\}-\{Zh\} or \{Zh\}-\{Ja\} translations.
We choose these three languages because English has the largest share of training data~\cite{brown2020language, chowdhery2022palm}.
Japanese, as a typologically distinct language, exhibits substantial orthographic and structural differences from English~\cite{haspelmath2005world, deguchi-iclr-2025-softmatcha}.
Meanwhile, Chinese, with a data volume comparable to English, is also included for this analysis.
For each language pair, we create 2,000 samples, and for the geo-culture task, we prepare 2,000 samples for each language as well.

\subsection{Adversarial Prompts Settings}
\label{adversarial-prompts-settings}
As described in Section~\ref{adversarial-prompts}, we attempt to disrupt consistency by injecting adversarial prompts.
We systematically examine the proportion of the input length occupied by these adversarial prompts, as detailed in Appendix~\ref{app:details-of-input-token-length}, using ratios of 20\%, 40\%, 60\%, 80\%, and 100\% of the model’s maximum input token length (e.g., 32,768 tokens for Qwen2.5).
By varying these ratios incrementally, our work makes it possible to systematically analyze the impact of adversarial prompts.
The input languages include Japanese, Chinese, and English.
Appendix~\ref{app:detailed-prompts} provides examples of the prompts, and Table~\ref{tab:experimental-sesttins} shows all experimental settings.
    
\subsection{Templates}
\label{templates}
The model input consists of the problem statements created in Section~\ref{dataset-creation}.
Following prior work~\cite{wendler-etal-2024-llamas}, we conduct experiments using 4-shot to improve the model's ability to accurately fill in the blanks.


\section{Results}
\label{results}
\begin{figure*}[t]
  \begin{minipage}[b]{0.5\linewidth}
    \centering
    \includegraphics[width=\linewidth]{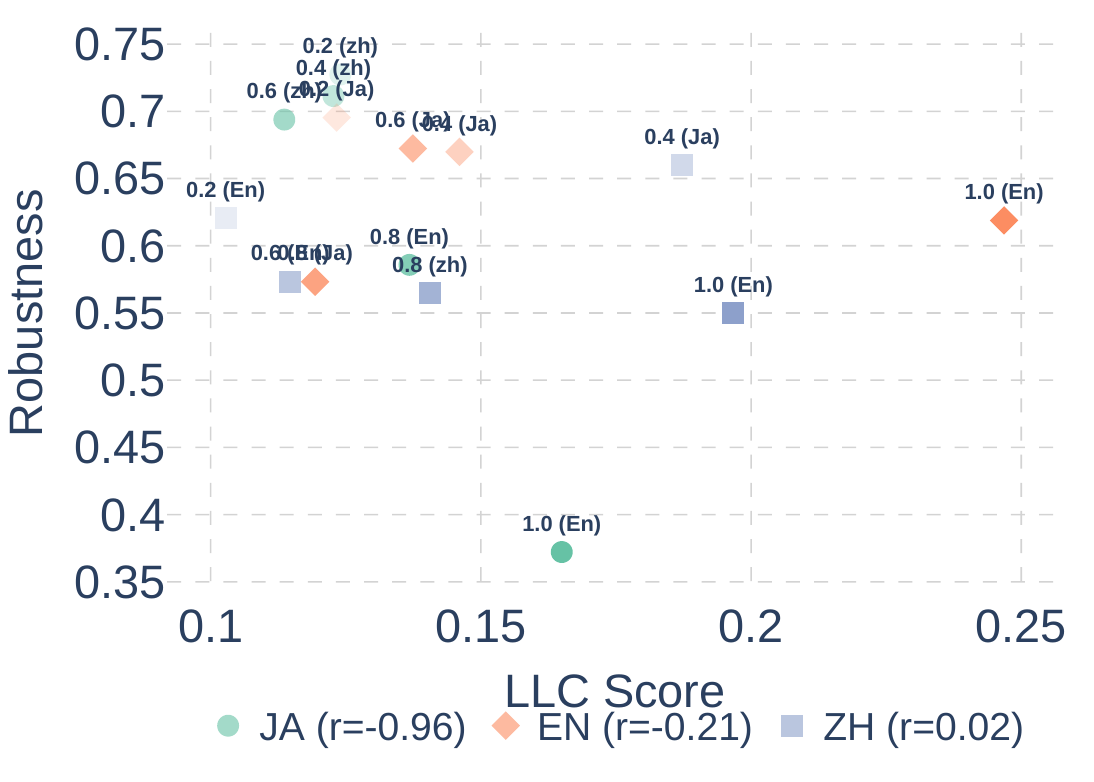}
    \caption*{(a) Src: En, Trg: Ja}
  \end{minipage}
  \begin{minipage}[b]{0.5\linewidth}
    \centering
    \includegraphics[width=\linewidth]{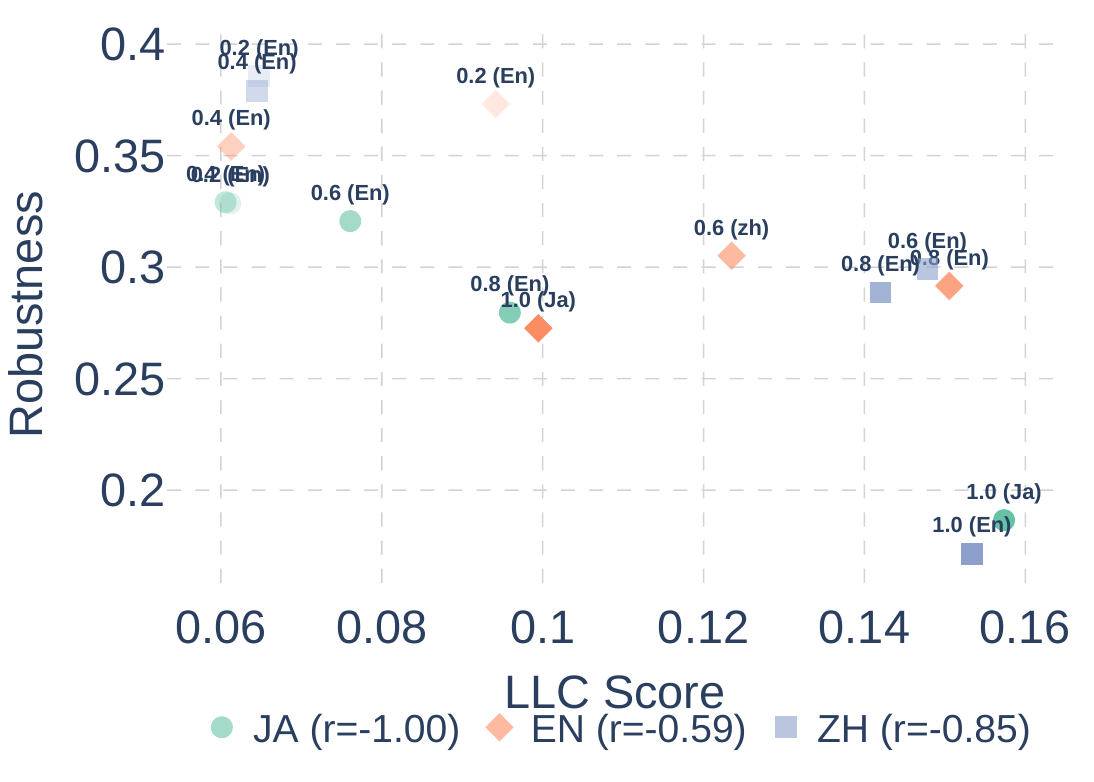}
    \caption*{(b) Src: En, Trg: Zh}
  \end{minipage}
\caption{
Correlation (\(r\)) between language consistency and robustness in translation tasks using LLM-jp-3.
For example, $\overset{\text{Ratio (Lang)}}{\textcolor{green!50!black}{\text{\small\ding{108}}}}$ indicates when inserting Japanese adversarial prompts with (Ratio)\%, the model thinks in the latent language (Lang).
}
\label{correlation-translation-llm-jp1}
\end{figure*}

\begin{figure*}[t]
  \begin{minipage}[b]{0.5\linewidth}
    \centering
    \includegraphics[width=\linewidth]{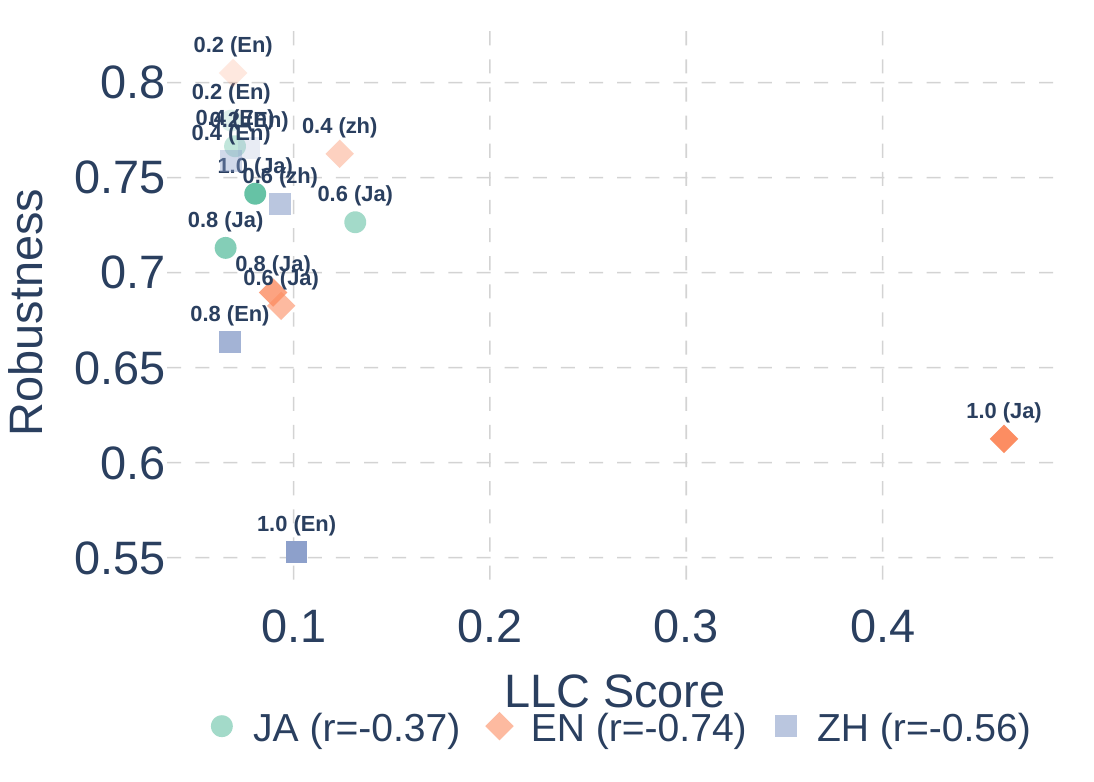}
    \caption*{(a) Src: Ja, Trg: En}
  \end{minipage}
  \begin{minipage}[b]{0.5\linewidth}
    \centering
    \includegraphics[width=\linewidth]{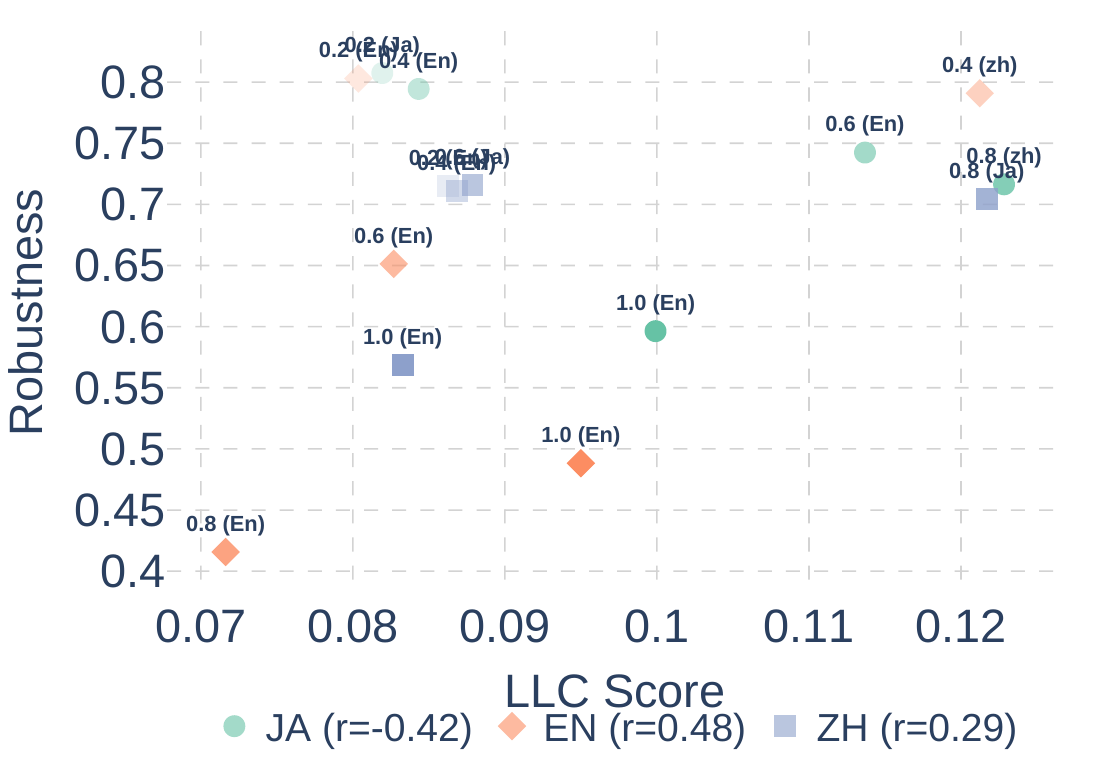}
    \caption*{(b) Src: Zh, Trg: En}
  \end{minipage}
\caption{
Same as Figure~\ref{correlation-translation-llm-jp1}, but (a) is \{Ja\}-\{En\} and (b) is \{Zh\}-\{En\}.
}
  \label{correlation-translation-llm-jp2}
\end{figure*}

\begin{figure*}[t]
  \begin{minipage}[b]{0.5\linewidth}
    \centering
    \includegraphics[width=\linewidth]{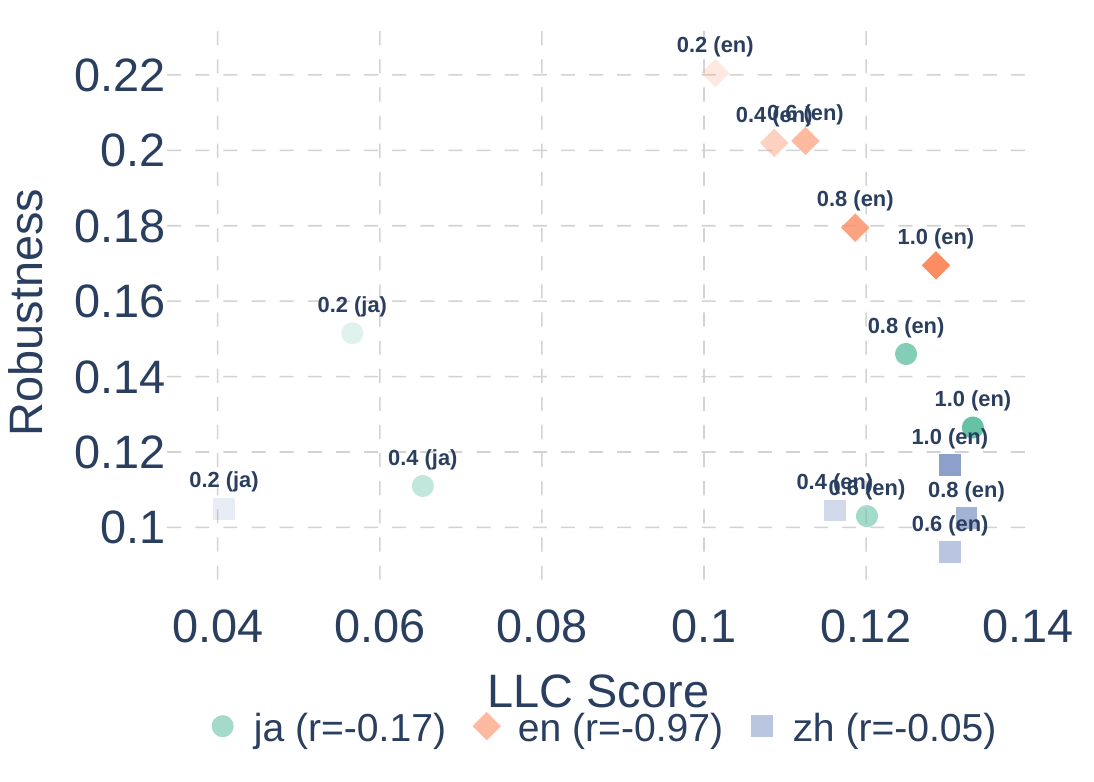}
    \caption*{(a) Question in English}
  \end{minipage}
  \begin{minipage}[b]{0.5\linewidth}
    \centering
    \includegraphics[width=\linewidth]{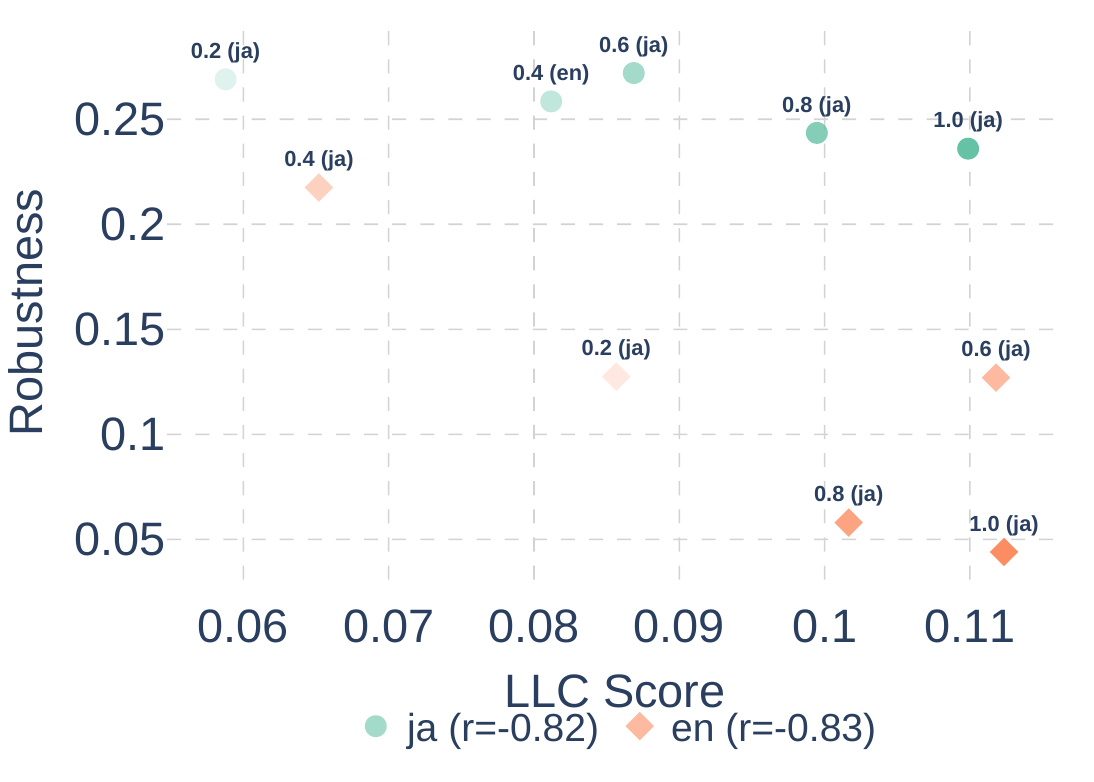}
    \caption*{(b) Question in Japanese}
  \end{minipage}
\caption{
Correlation (\(r\)) between consistency and robustness in geo-culture tasks using LLM-jp-3.
For example, $\overset{\text{Ratio (Lang)}}{\textcolor{orange}{\text{\small\ding{117}}}}$ indicates when inserting English adversarial prompts with (Ratio)\%, the model thinks in the latent language (Lang).
}
  \label{correlation-geo-llm-jp}
\end{figure*}

\begin{figure*}[t]
  \begin{minipage}[b]{0.5\linewidth}
    \centering
    \includegraphics[width=\linewidth]{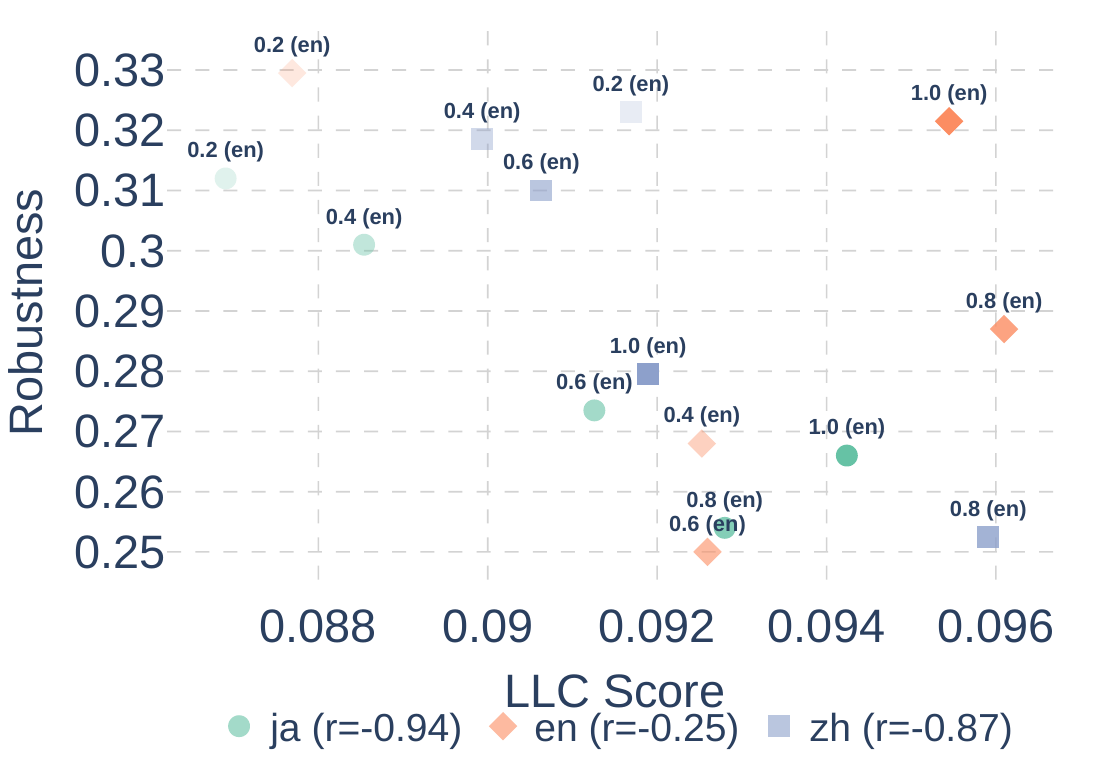}
    \caption*{(a) Question in English}
  \end{minipage}
  \begin{minipage}[b]{0.5\linewidth}
    \centering
    \includegraphics[width=\linewidth]{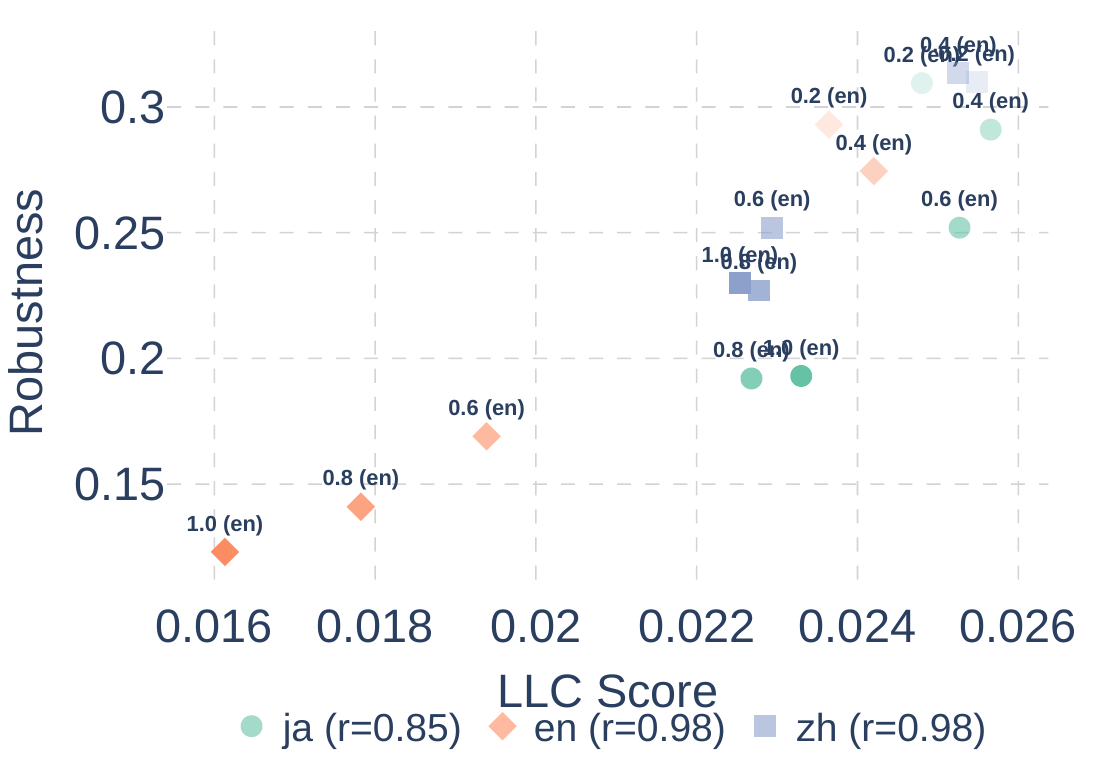}
    \caption*{(b) Question in English}
  \end{minipage}
\caption{
Same as Figure~\ref{correlation-translation-llm-jp1}, but with different models. (a) shows Qwen2.5 and (b) shows Gemma3.
}
  \label{correlation-geo-qwen-and-gemma}
\end{figure*}

Figures~\ref{correlation-translation-llm-jp1} and~\ref{correlation-translation-llm-jp2} present the experimental results for the translation task, while Figures~\ref{correlation-geo-llm-jp} and~\ref{correlation-geo-qwen-and-gemma} show the results for the geo-culture task.
In these plots, the x-axis represents the LLC Score, and the y-axis represents the robustness.
Each point represents the result on a specific task.
$\overset{\text{1.0 (Ja)}}{\textcolor{orange}{\text{\ding{117}}}}$ indicates the result when adversarial prompts in English make up 100\% of the input, resulting in the latent language being Japanese (Ja).
We denote the correlation between Robustness and the LLC Score as $r$.
If our hypothesis is correct, the data points should align along the line $y = -x$, indicating that LLC Score should lead to a corresponding drop in robustness.

In Figure~\ref{correlation-translation-llm-jp1} (b), when looking at the results for Chinese adversarial prompts, we observe that lower ratios of adversarial prompts, e.g., $\overset{\text{0.2 (En)}}{\textcolor{violet}{\text{\small\ding{110}}}}$ or $\overset{\text{0.4 (En)}}{\textcolor{violet}{\text{\small\ding{110}}}}$, lead to higher Robustness and lower LLC Score.
As the ratio increases, both metrics degrade, indicating a negative correlation ($r = -0.85$).
Similar behavior appears with adversarial prompts in other languages as well ($r = -1.0$ for Japanese, and $r = -0.59$ for English).
In contrast, Figures~\ref{correlation-translation-llm-jp2} and~\ref{correlation-geo-qwen-and-gemma} show different trends, i.e., ($r = 0.29$ for Chinese in Figure~\ref{correlation-translation-llm-jp2}, and $r = 0.98$ for English and Chinse in Figure~\ref{correlation-geo-qwen-and-gemma}), making the overall pattern less clear.
Looking at each point, we observe that the latent language remains unchanged regardless of whether a correlation exists or not.
This observation suggests that models do not necessarily need to operate in their proficient latent language to perform downstream tasks effectively.
These results contradict the initial hypothesis that disrupting thought in latent language affects downstream task performance.

\section{Discussion}
\subsection{Correlation Between Robustness and Latent Language}
Figures~\ref{correlation-translation-llm-jp1}, \ref{correlation-translation-llm-jp2}, and \ref{correlation-geo-llm-jp} illustrate the correlation between consistency and robustness in translation and geo-culture tasks.
The results tend to align with the ideal line $y = -x$, indicating that a loss of robustness is often accompanied by a disruption in consistency.
For example, in Figure~\ref{correlation-geo-llm-jp}, when the adversarial prompts are in English, the points align well along the line $y = -x$.

Focusing on the y-axis, we observe that when the question language or the source language matches the adversarial prompt, the model tends to maintain higher robustness.
In Figure~\ref{correlation-geo-llm-jp} (b), where the question is in English, the highest accuracy, i.e., the highest robustness, is achieved when the adversarial prompt is also in English.
We can observe a similar tendency in (a) as well.

In contrast, adversarial prompts in different languages generally lead to lower robustness.
This suggests that mismatches between the question language and the adversarial prompt can largely disrupt downstream task performance.
However, examining the x-axis reveals a different pattern: even when the adversarial prompt language differs from the question language, the expected consistency breakdown does not always occur.
This finding indicates that models can flexibly adjust to different latent languages depending on the input language.

\subsection{Impact of Adversarial Prompt Ratio on Consistency}
We analyze the impact of adversarial prompt ratio on consistency.
In many cases, increasing this ratio of adversarial prompts clearly affects internal consistency as shown in Figures~\ref{correlation-translation-llm-jp1}, \ref{correlation-translation-llm-jp2}, and \ref{correlation-geo-llm-jp}.
However, when the adversarial prompt language differs from the question language, the latent language processing appears unaffected.
For example, in Figure~\ref{correlation-geo-llm-jp} (a), since the question is in English, we would ideally expect the LLC Score to be low when the latent language is English. 
However, the results show that the latent language is Japanese instead.
This result implies that consistency in latent language is resilient to adversarial prompts, supporting the view that models do not necessarily need to operate in their preferred latent language to maintain internal consistency.

\subsection{Impact of Adversarial Prompt Ratio on Robustness}
Increasing the proportion of adversarial prompts largely impacts model robustness across most models.
This observation aligns with prior studies~\cite{goel-etal-2021-robustness, khashabi-etal-2020-bang}, which have shown that increasing noise in the input degrades downstream task performance.

\subsection{Data Distribution Analysis}
\label{data-distribution-analysis}
\begin{figure*}[t]
  \begin{minipage}[b]{0.49\linewidth}
    \centering
    \includegraphics[width=\linewidth]{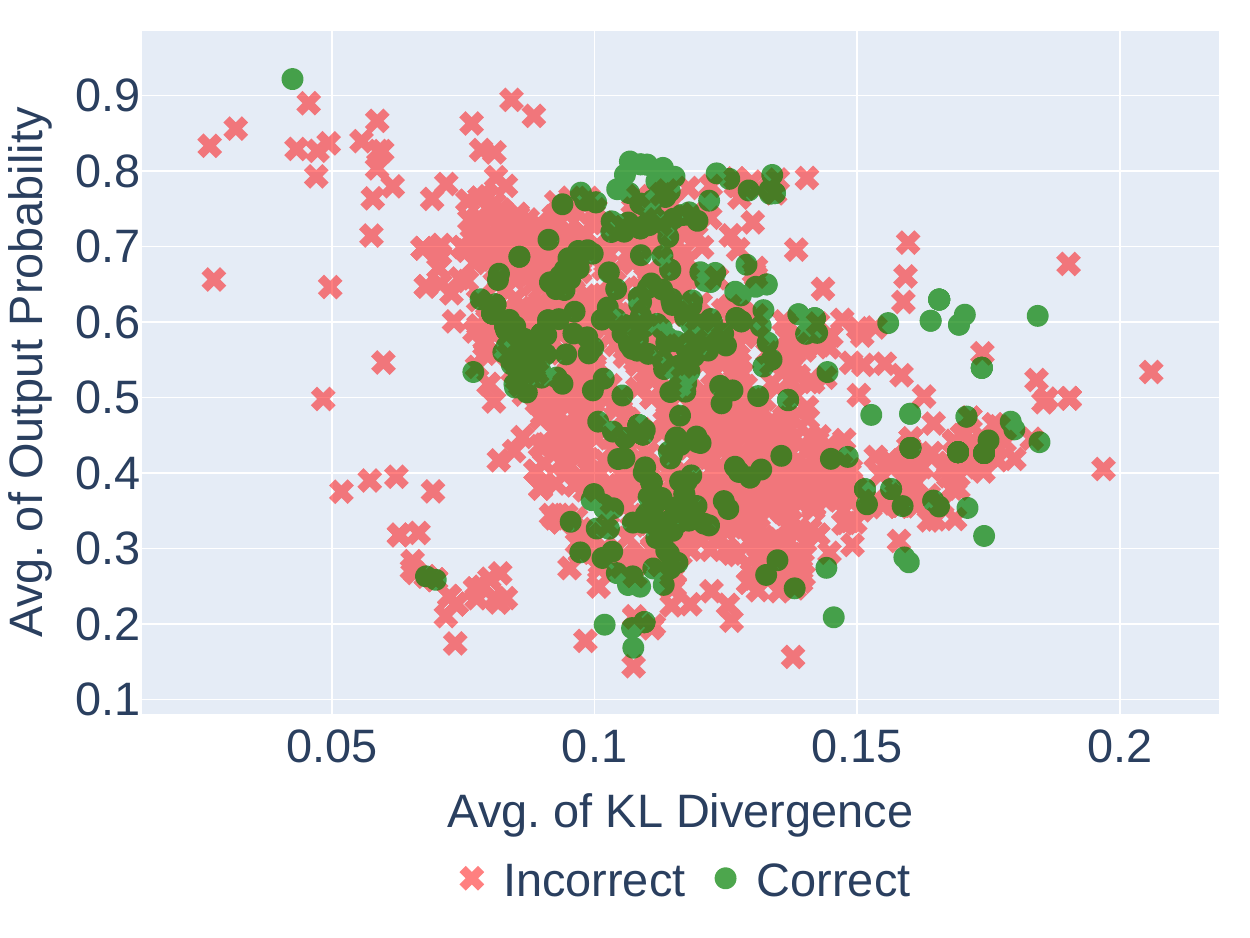}
    \caption*{(a) Question in Japanese w/o adversarial}
  \end{minipage}
  \begin{minipage}[b]{0.49\linewidth}
    \centering
    \includegraphics[width=\linewidth]{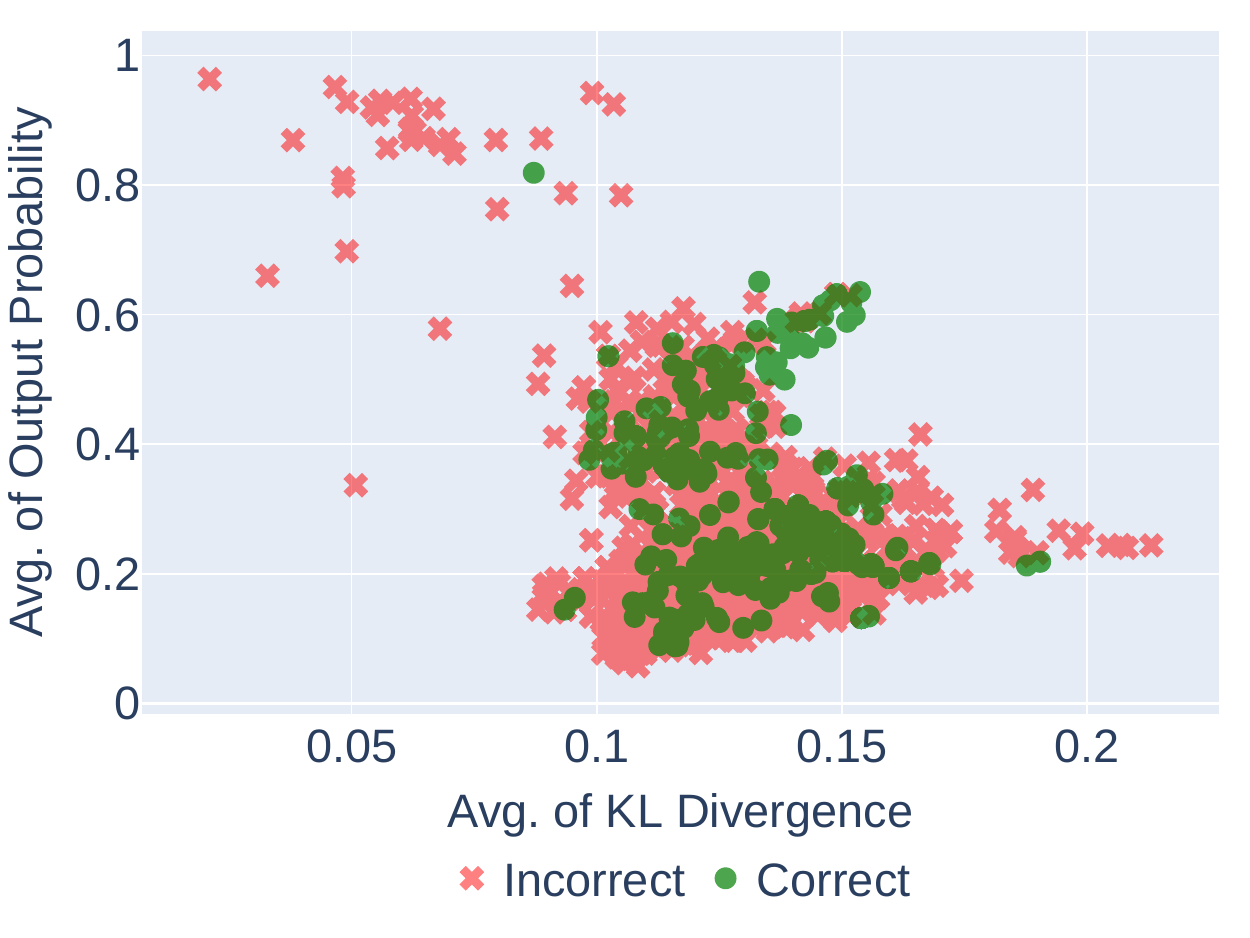}
    \caption*{(b) Question and adversarial in Japanese}
  \end{minipage}
\caption{
Results of the LLM-jp-3 model on the geo-culture task. The left panel shows results without noise, while the right panel shows results with 100\% noise.
Out of 2,000 samples, green points indicate correct predictions, while red points indicate incorrect ones.
}
  \label{data-plot-geo}
\end{figure*}

To investigate how latent language disruptions relate to internal consistency  and model confidence, we analyze the correlation between KL divergence in each adjacent layer and output probabilities at the question level.
Figure~\ref{data-plot-geo} presents a scatter plot of KL values and output probabilities for each question in the geo-culture task.
The x-axis represents KL divergence values (lower is better), while the y-axis shows output probabilities (higher is better).
The KL values are computed for each layer using the LogitLens, calculating the KL divergence between consecutive layers (\(KL_{l,l+1}\) in Section~\ref{analysis-method}).
Green circles (\textbf{\textcolor{green!50!black}{\ding{108}}}) indicate correct responses, while red crosses (\textbf{\textcolor{red!50!black}{$\times$}}) represent incorrect ones.
Figure (a) includes no adversarial prompts, while (b) includes 100\% adversarial prompts.
According to Table~\ref{tab:result-table-geo}, the accuracy for these two settings is 0.25 in left and 0.24 in right respectively.
This analysis demonstrates that inserting adversarial prompts reduces model confidence and robustness, as indicated by the overall downward shift in output probabilities.

\section{Conclusion}
\label{conclusion}
In our study, we systematically analyze the impact of consistency in latent language on downstream task performance.
Specifically, we focus on understanding how disruptions in consistency of the latent language influence their reasoning processes and overall task performance.
To achieve this, we insert adversarial prompts containing multiple languages, deliberately disrupting the models' internal representations and quantitatively measuring the resulting performance shifts.

Our findings indicate that, contrary to our initial hypothesis, maintaining a consistent latent language is not always necessary for achieving best task performance.
Models can flexibly adapt to varying input languages near the final layers, suggesting that their internal reasoning processes are more resilient to linguistic perturbations than previously thought.
However, tasks may be heavily dependent on precise linguistic alignment, such as translation, are particularly sensitive to disruptions by adversarial prompts.
In these cases, the introduction of adversarial prompts from less familiar languages largely degrades performance, likely due to destabilization of the models' internal structures.

\clearpage
\bibliography{custom, anthology}

\begin{thebibliography}{10}

\bibitem{wendler-etal-2024-llamas}
Chris Wendler, Veniamin Veselovsky, Giovanni Monea, and Robert West.
\newblock Do llamas work in {E}nglish? on the latent language of multilingual transformers.
\newblock In Lun-Wei Ku, Andre Martins, and Vivek Srikumar, editors, {\em Proceedings of the 62nd Annual Meeting of the Association for Computational Linguistics (Volume 1: Long Papers)}, pages 15366--15394, Bangkok, Thailand, August 2024. Association for Computational Linguistics.

\bibitem{zhong2024beyond}
Chengzhi Zhong, Fei Cheng, Qianying Liu, Junfeng Jiang, Zhen Wan, Chenhui Chu, Yugo Murawaki, and Sadao Kurohashi.
\newblock Beyond english-centric llms: What language do multilingual language models think in?
\newblock {\em arXiv preprint arXiv:2408.10811}, 2024.

\bibitem{touvron2023llama}
Hugo Touvron, Louis Martin, Kevin Stone, Peter Albert, Amjad Almahairi, Yasmine Babaei, Nikolay Bashlykov, Soumya Batra, Prajjwal Bhargava, Shruti Bhosale, et~al.
\newblock Llama 2: Open foundation and fine-tuned chat models.
\newblock {\em arXiv preprint arXiv:2307.09288}, 2023.

\bibitem{aizawa2024llm}
Akiko Aizawa, Eiji Aramaki, Bowen Chen, Fei Cheng, Hiroyuki Deguchi, Rintaro Enomoto, Kazuki Fujii, Kensuke Fukumoto, Takuya Fukushima, Namgi Han, et~al.
\newblock Llm-jp: A cross-organizational project for the research and development of fully open japanese llms.
\newblock {\em arXiv preprint arXiv:2407.03963}, 2024.

\bibitem{yong2025crosslingualreasoningtesttimescaling}
Zheng-Xin Yong, M.~Farid Adilazuarda, Jonibek Mansurov, Ruochen Zhang, Niklas Muennighoff, Carsten Eickhoff, Genta~Indra Winata, Julia Kreutzer, Stephen~H. Bach, and Alham~Fikri Aji.
\newblock Crosslingual reasoning through test-time scaling, 2025.

\bibitem{sakai-etal-2024-mcsqa}
Yusuke Sakai, Hidetaka Kamigaito, and Taro Watanabe.
\newblock m{CSQA}: Multilingual commonsense reasoning dataset with unified creation strategy by language models and humans.
\newblock In Lun-Wei Ku, Andre Martins, and Vivek Srikumar, editors, {\em Findings of the Association for Computational Linguistics: ACL 2024}, pages 14182--14214, Bangkok, Thailand, August 2024. Association for Computational Linguistics.

\bibitem{ozaki2024bqa}
Shintaro Ozaki, Kazuki Hayashi, Miyu Oba, Yusuke Sakai, Hidetaka Kamigaito, and Taro Watanabe.
\newblock Bqa: Body language question answering dataset for video large language models.
\newblock {\em arXiv preprint arXiv:2410.13206}, 2024.

\bibitem{schut2025multilingual}
Lisa Schut, Yarin Gal, and Sebastian Farquhar.
\newblock Do multilingual llms think in english?
\newblock {\em arXiv preprint arXiv:2502.15603}, 2025.

\bibitem{haemmerl-etal-2023-speaking}
Katharina H{\"a}mmerl, Bjoern Deiseroth, Patrick Schramowski, Jind{\v{r}}ich Libovick{\'y}, Constantin Rothkopf, Alexander Fraser, and Kristian Kersting.
\newblock Speaking multiple languages affects the moral bias of language models.
\newblock In Anna Rogers, Jordan Boyd-Graber, and Naoaki Okazaki, editors, {\em Findings of the Association for Computational Linguistics: ACL 2023}, pages 2137--2156, Toronto, Canada, July 2023. Association for Computational Linguistics.

\bibitem{gallegos2023bias}
IO~Gallegos, RA~Rossi, J~Barrow, MM~Tanjim, S~Kim, F~Dernoncourt, T~Yu, R~Zhang, and NK~Ahmed.
\newblock Bias and fairness in large language models: A survey. arxiv, 2023.

\bibitem{huang-etal-2023-languages}
Haoyang Huang, Tianyi Tang, Dongdong Zhang, Xin Zhao, Ting Song, Yan Xia, and Furu Wei.
\newblock Not all languages are created equal in {LLM}s: Improving multilingual capability by cross-lingual-thought prompting.
\newblock In Houda Bouamor, Juan Pino, and Kalika Bali, editors, {\em Findings of the Association for Computational Linguistics: EMNLP 2023}, pages 12365--12394, Singapore, December 2023. Association for Computational Linguistics.

\bibitem{shi2022language}
Freda Shi, Mirac Suzgun, Markus Freitag, Xuezhi Wang, Suraj Srivats, Soroush Vosoughi, Hyung~Won Chung, Yi~Tay, Sebastian Ruder, Denny Zhou, et~al.
\newblock Language models are multilingual chain-of-thought reasoners.
\newblock {\em arXiv preprint arXiv:2210.03057}, 2022.

\bibitem{lai-nissim-2024-mcot}
Huiyuan Lai and Malvina Nissim.
\newblock m{C}o{T}: Multilingual instruction tuning for reasoning consistency in language models.
\newblock In Lun-Wei Ku, Andre Martins, and Vivek Srikumar, editors, {\em Proceedings of the 62nd Annual Meeting of the Association for Computational Linguistics (Volume 1: Long Papers)}, pages 12012--12026, Bangkok, Thailand, August 2024. Association for Computational Linguistics.

\bibitem{wei2022chain}
Jason Wei, Xuezhi Wang, Dale Schuurmans, Maarten Bosma, Fei Xia, Ed~Chi, Quoc~V Le, Denny Zhou, et~al.
\newblock Chain-of-thought prompting elicits reasoning in large language models.
\newblock {\em Advances in neural information processing systems}, 35:24824--24837, 2022.

\bibitem{xue-etal-2021-mt5}
Linting Xue, Noah Constant, Adam Roberts, Mihir Kale, Rami Al-Rfou, Aditya Siddhant, Aditya Barua, and Colin Raffel.
\newblock m{T}5: A massively multilingual pre-trained text-to-text transformer.
\newblock In Kristina Toutanova, Anna Rumshisky, Luke Zettlemoyer, Dilek Hakkani-Tur, Iz~Beltagy, Steven Bethard, Ryan Cotterell, Tanmoy Chakraborty, and Yichao Zhou, editors, {\em Proceedings of the 2021 Conference of the North American Chapter of the Association for Computational Linguistics: Human Language Technologies}, pages 483--498, Online, June 2021. Association for Computational Linguistics.

\bibitem{goel-etal-2021-robustness}
Karan Goel, Nazneen~Fatema Rajani, Jesse Vig, Zachary Taschdjian, Mohit Bansal, and Christopher R{\'e}.
\newblock Robustness gym: Unifying the {NLP} evaluation landscape.
\newblock In Avi Sil and Xi~Victoria Lin, editors, {\em Proceedings of the 2021 Conference of the North American Chapter of the Association for Computational Linguistics: Human Language Technologies: Demonstrations}, pages 42--55, Online, June 2021. Association for Computational Linguistics.

\bibitem{khashabi-etal-2020-bang}
Daniel Khashabi, Tushar Khot, and Ashish Sabharwal.
\newblock More bang for your buck: Natural perturbation for robust question answering.
\newblock In Bonnie Webber, Trevor Cohn, Yulan He, and Yang Liu, editors, {\em Proceedings of the 2020 Conference on Empirical Methods in Natural Language Processing (EMNLP)}, pages 163--170, Online, November 2020. Association for Computational Linguistics.

\bibitem{nostalgebraist2020logitlens}
nostalgebraist.
\newblock Interpreting gpt: the logit lens.
\newblock \url{https://www.lesswrong.com/posts/AcKRB8wDpdaN6v6ru/interpreting-gpt-the-logit-lens}, August 2020.
\newblock LessWrong.

\bibitem{tunedlens}
Nora Belrose, Zach Furman, Logan Smith, Danny Halawi, Igor Ostrovsky, Lev McKinney, Stella Biderman, and Jacob Steinhardt.
\newblock Eliciting latent predictions from transformers with the tuned lens, 2023.

\bibitem{vaswani2017attention}
Ashish Vaswani, Noam Shazeer, Niki Parmar, Jakob Uszkoreit, Llion Jones, Aidan~N Gomez, {\L}ukasz Kaiser, and Illia Polosukhin.
\newblock Attention is all you need.
\newblock {\em Advances in neural information processing systems}, 30, 2017.

\bibitem{belrose2023eliciting}
Nora Belrose, Zach Furman, Logan Smith, Danny Halawi, Igor Ostrovsky, Lev McKinney, Stella Biderman, and Jacob Steinhardt.
\newblock Eliciting latent predictions from transformers with the tuned lens.
\newblock {\em arXiv preprint arXiv:2303.08112}, 2023.

\bibitem{petroni-etal-2019-language}
Fabio Petroni, Tim Rockt{\"a}schel, Sebastian Riedel, Patrick Lewis, Anton Bakhtin, Yuxiang Wu, and Alexander Miller.
\newblock Language models as knowledge bases?
\newblock In Kentaro Inui, Jing Jiang, Vincent Ng, and Xiaojun Wan, editors, {\em Proceedings of the 2019 Conference on Empirical Methods in Natural Language Processing and the 9th International Joint Conference on Natural Language Processing (EMNLP-IJCNLP)}, pages 2463--2473, Hong Kong, China, November 2019. Association for Computational Linguistics.

\bibitem{taylor1953cloze}
Wilson~L Taylor.
\newblock “cloze procedure”: A new tool for measuring readability.
\newblock {\em Journalism quarterly}, 30(4):415--433, 1953.

\bibitem{samuel2024personagym}
Vinay Samuel, Henry~Peng Zou, Yue Zhou, Shreyas Chaudhari, Ashwin Kalyan, Tanmay Rajpurohit, Ameet Deshpande, Karthik Narasimhan, and Vishvak Murahari.
\newblock Personagym: Evaluating persona agents and llms.
\newblock {\em arXiv preprint arXiv:2407.18416}, 2024.

\bibitem{zhu2018texygen}
Yaoming Zhu, Sidi Lu, Lei Zheng, Jiaxian Guo, Weinan Zhang, Jun Wang, and Yong Yu.
\newblock Texygen: A benchmarking platform for text generation models.
\newblock In {\em The 41st international ACM SIGIR conference on research \& development in information retrieval}, pages 1097--1100, 2018.

\bibitem{hurst2024gpt}
Aaron Hurst, Adam Lerer, Adam~P Goucher, Adam Perelman, Aditya Ramesh, Aidan Clark, AJ~Ostrow, Akila Welihinda, Alan Hayes, Alec Radford, et~al.
\newblock Gpt-4o system card.
\newblock {\em arXiv preprint arXiv:2410.21276}, 2024.

\bibitem{team2025gemma}
Gemma Team, Aishwarya Kamath, Johan Ferret, Shreya Pathak, Nino Vieillard, Ramona Merhej, Sarah Perrin, Tatiana Matejovicova, Alexandre Ram{\'e}, Morgane Rivi{\`e}re, et~al.
\newblock Gemma 3 technical report.
\newblock {\em arXiv preprint arXiv:2503.19786}, 2025.

\bibitem{yang2024qwen2}
An~Yang, Baosong Yang, Beichen Zhang, Binyuan Hui, Bo~Zheng, Bowen Yu, Chengyuan Li, Dayiheng Liu, Fei Huang, Haoran Wei, et~al.
\newblock Qwen2. 5 technical report.
\newblock {\em arXiv preprint arXiv:2412.15115}, 2024.

\bibitem{brown2020language}
Tom Brown, Benjamin Mann, Nick Ryder, Melanie Subbiah, Jared~D Kaplan, Prafulla Dhariwal, Arvind Neelakantan, Pranav Shyam, Girish Sastry, Amanda Askell, et~al.
\newblock Language models are few-shot learners.
\newblock {\em Advances in neural information processing systems}, 33:1877--1901, 2020.

\bibitem{chowdhery2022palm}
Aakanksha Chowdhery, Sharan Narang, Jacob Devlin, Maarten Bosma, Gaurav Mishra, Adam Roberts, Paul Barham, Hyung~Won Chung, Charles Sutton, Sebastian Gehrmann, et~al.
\newblock Palm: Scaling language modeling with pathways. arxiv 2022.
\newblock {\em arXiv preprint arXiv:2204.02311}, 10:1, 2022.

\bibitem{haspelmath2005world}
M.~Haspelmath, M.S. Dryer, D.~Gil, and B.~Comrie.
\newblock {\em The World Atlas of Language Structures}.
\newblock Number v. 1 in Oxford linguistics. OUP Oxford, 2005.

\bibitem{deguchi-iclr-2025-softmatcha}
Hiroyuki Deguchi, Go~Kamoda, Yusuke Matsushita, Chihiro Taguchi, Masaki Waga, Kohei Suenaga, and Sho Yokoi.
\newblock Softmatcha: A soft and fast pattern matcher for billion-scale corpus searches.
\newblock In {\em The Thirteenth International Conference on Learning Representations (ICLR 2025)}, 2025.

\bibitem{wolf-etal-2020-transformers}
Thomas Wolf, Lysandre Debut, Victor Sanh, Julien Chaumond, Clement Delangue, Anthony Moi, Pierric Cistac, Tim Rault, Remi Louf, Morgan Funtowicz, Joe Davison, Sam Shleifer, Patrick von Platen, Clara Ma, Yacine Jernite, Julien Plu, Canwen Xu, Teven Le~Scao, Sylvain Gugger, Mariama Drame, Quentin Lhoest, and Alexander Rush.
\newblock Transformers: State-of-the-art natural language processing.
\newblock In Qun Liu and David Schlangen, editors, {\em Proceedings of the 2020 Conference on Empirical Methods in Natural Language Processing: System Demonstrations}, pages 38--45, Online, October 2020. Association for Computational Linguistics.

\bibitem{dettmers2022gpt3}
Tim Dettmers, Mike Lewis, Younes Belkada, and Luke Zettlemoyer.
\newblock Gpt3. int8 (): 8-bit matrix multiplication for transformers at scale.
\newblock {\em Advances in Neural Information Processing Systems}, 35:30318--30332, 2022.

\end{thebibliography}
\bibliographystyle{unsrt}

\clearpage
\section*{NeurIPS Paper Checklist}

\begin{enumerate}

\item {\bf Claims}
    \item[] Question: Do the main claims made in the abstract and introduction accurately reflect the paper's contributions and scope?
    \item[] Answer: \answerYes{} 
    \item[] Justification: We stated our contributions and scope of in the abstract and introduction, and the statements are supported by the experimental results.

\item {\bf Limitations}
    \item[] Question: Does the paper discuss the limitations of the work performed by the authors?
    \item[] Answer: \answerYes{} 
    \item[] Justification: We discussed our limitations in Appendix~\ref{limitations}.

\item {\bf Theory assumptions and proofs}
    \item[] Question: For each theoretical result, does the paper provide the full set of assumptions and a complete (and correct) proof?
    \item[] Answer: \answerNA{} 
    \item[] Justification: This is an interpretation paper without any theoretical results.

    \item {\bf Experimental result reproducibility}
    \item[] Question: Does the paper fully disclose all the information needed to reproduce the main experimental results of the paper to the extent that it affects the main claims and/or conclusions of the paper (regardless of whether the code and data are provided or not)?
    \item[] Answer: \answerYes{} 
    \item[] Justification: We provided detailed model settings in Appendix~\ref{app:detailed-model-settings}, detailed model name we used in Appendix~\ref{tab:detailed-model-names}, and detailed prompts which we instructed in Appendix~\ref{app:detailed-prompts}.

\item {\bf Open access to data and code}
    \item[] Question: Does the paper provide open access to the data and code, with sufficient instructions to faithfully reproduce the main experimental results, as described in supplemental material?
    \item[] Answer: \answerNA{} 
    \item[] Justification: As we mentioned in Appendix~\ref{app:detailed-model-settings}, due to double blind anonymized policy, we will release upon acceptance.

\item {\bf Experimental setting/details}
    \item[] Question: Does the paper specify all the training and test details (e.g., data splits, hyperparameters, how they were chosen, type of optimizer, etc.) necessary to understand the results?
    \item[] Answer: \answerYes{} 
    \item[] Justification: We mentioned in Section~\ref{dataset-creation}, Section~\ref{models}, Appendix~\ref{app:detailed-model-settings}, and Appendix~\ref{tab:detailed-model-names}.

\item {\bf Experiment statistical significance}
    \item[] Question: Does the paper report error bars suitably and correctly defined or other appropriate information about the statistical significance of the experiments?
    \item[] Answer: \answerNo{} 
    \item[] Justification: We do not report the error bars for the measurement of effectivity and retainability, as it is required to run the pretraining experiments multiple times with different seeds, which is not feasible due to the restriction of computational resources.

\item {\bf Experiments compute resources}
    \item[] Question: For each experiment, does the paper provide sufficient information on the computer resources (type of compute workers, memory, time of execution) needed to reproduce the experiments?
    \item[] Answer: \answerYes{} 
    \item[] Justification: We mentioned in Appendix~\ref{app:inference-settings}.
    
\item {\bf Code of ethics}
    \item[] Question: Does the research conducted in the paper conform, in every respect, with the NeurIPS Code of Ethics \url{https://neurips.cc/public/EthicsGuidelines}?
    \item[] Answer: \answerYes{} 
    \item[] Justification:  We have confirmed that this work is done following the NeurIPS Code of Ethics.

\item {\bf Broader impacts}
    \item[] Question: Does the paper discuss both potential positive societal impacts and negative societal impacts of the work performed?
    \item[] Answer: \answerYes{} 
    \item[] Justification: We discussed our impact in Abstract, Introduction, and Conclusion.
    
\item {\bf Safeguards}
    \item[] Question: Does the paper describe safeguards that have been put in place for responsible release of data or models that have a high risk for misuse (e.g., pretrained language models, image generators, or scraped datasets)?
    \item[] Answer: \answerNA{} 
    \item[] Justification:  Our work poses no risk for misuse.

\item {\bf Licenses for existing assets}
    \item[] Question: Are the creators or original owners of assets (e.g., code, data, models), used in the paper, properly credited and are the license and terms of use explicitly mentioned and properly respected?
    \item[] Answer: \answerYes{} 
    \item[] Justification: We cited and explicitly mentioned the original papers that produced the pretrained models or datasets that are used in this work.

\item {\bf New assets}
    \item[] Question: Are new assets introduced in the paper well documented and is the documentation provided alongside the assets?
    \item[] Answer: \answerYes{} 
    \item[] Justification: We mentioned carefully in Section~\ref{dataset-creation} and Appendix~\ref{app:detailed-prompts}.

\item {\bf Crowdsourcing and research with human subjects}
    \item[] Question: For crowdsourcing experiments and research with human subjects, does the paper include the full text of instructions given to participants and screenshots, if applicable, as well as details about compensation (if any)? 
    \item[] Answer: \answerNA{} 
    \item[] Justification: Our work does not involve crowdsourcing nor research with human subjects.

\item {\bf Institutional review board (IRB) approvals or equivalent for research with human subjects}
    \item[] Question: Does the paper describe potential risks incurred by study participants, whether such risks were disclosed to the subjects, and whether Institutional Review Board (IRB) approvals (or an equivalent approval/review based on the requirements of your country or institution) were obtained?
    \item[] Answer: \answerNA{} 
    \item[] Justification: Our work does not involve crowdsourcing nor research with human subjects.

\item {\bf Declaration of LLM usage}
    \item[] Question: Does the paper describe the usage of LLMs if it is an important, original, or non-standard component of the core methods in this research? Note that if the LLM is used only for writing, editing, or formatting purposes and does not impact the core methodology, scientific rigorousness, or originality of the research, declaration is not required.
    \item[] Answer: \answerYes{} 
    \item[] Justification: We mentioned in Appendix~\ref{app:ai-assistant-tools}.

\end{enumerate}

\appendix
 \section{Limitations}
\label{limitations}
Our study has several limitations. 
First, there are various approaches for determining the dominant language.
In our study, we defined the LLC Score and identified the language with the lowest score as the dominant language, treating it as the latent language.
However, other, potentially better, approaches may exist.
Our approach was selected because it showed the strongest correlation with other potential methods, such as averaging 
KL scores which we mentioned in ~\ref{data-distribution-analysis}.

Moreover, our study investigates the correlation between robustness and consistency through the use of adversarial prompts.
This raises the possibility that the observed effects may depend on the specific adversarial prompts used.

Lastly, it would be valuable to extend this analysis to larger models as we mentioned in Appendix~\ref{app:results-by-larger-llms} partially and datasets.

\section{Appnendix}
\subsection{Questions Diversity in Datatasets}
\label{app:questions-diversity-in-datasets}
Self-BLEU \cite{zhu2018texygen}is a metric commonly used to evaluate the diversity of generated text within a dataset.
It measures the extent to which each sample in a set is distinct from the others, providing insight into the range of expressions present in the generated outputs.
A lower Self-BLEU score indicates higher diversity, while a higher score suggests redundancy or repetition in the dataset.

The Self-BLEU score for a single question $q_i$ is calculated as:

$$
\text{Self-BLEU}(q_i) = \text{BLEU}(q_i, \mathcal{Q} \setminus \{q_i\})
$$

Here, $q_i$ represents a single question from the dataset, $\mathcal{Q}$ denotes the complete set of questions in the dataset, and $\mathcal{Q} \setminus \{q_i\}$ is the set of all questions excluding $q_i$.
The overall Self-BLEU for the entire dataset is then the average of the individual Self-BLEU scores:

$$
\text{Self-BLEU}(\mathcal{Q}) = \frac{1}{|\mathcal{Q}|} \sum_{q_i \in \mathcal{Q}} \text{Self-BLEU}(q_i)
$$

where $|\mathcal{Q}|$ is the total number of questions in the dataset.
This formulation effectively captures the overlap between questions, providing a quantitative measure of diversity.
A more diverse dataset will naturally yield lower Self-BLEU scores, reflecting the presence of a broader range of expressions and phrasing within the question set.

To ensure more diverse problem statements generated by GPT-4, a category list was created, from which approximately 20 more specific categories were randomly selected to increase variability. The results are presented in Table~\ref{tab:self-bleu}.

\begin{tcolorbox}[
    title=Categories,
    boxrule=1pt,
    colback=white,
    coltitle=black,
    colframe=black,
    colbacktitle=white,
    arc=0mm, 
    outer arc=0mm,
    fonttitle=\bfseries,
    width=\textwidth,
    left=5pt,
    right=5pt,
    top=5pt,
    bottom=5pt
]
Capital City, Official Language, National Currency, Head of State, Independence Day, Major Religion, National Anthem, Famous Landmark, National Dish, Traditional Clothing, Flag Colors, Historical Figure, Major River, Mountain Range, Neighboring Countries, Climate, Population, Area (km²), UNESCO World Heritage Site, Government Type
\end{tcolorbox}

\begin{table}[h!]
\centering
\small
\caption{A result of Self-BLEU by 4-gram. This result shows that our created datasets have more diverse than preliminary ones which we ask GPT-4o to generate questions by same prompts.}
\begin{tabular}{lcccc}
\toprule
\multicolumn{5}{c}{\textbf{Self-BLEU ($\downarrow$)}} \\
\midrule
\multirow{2.5}{*}{\textbf{Task}} & \multicolumn{2}{c}{\textbf{Preliminary}} & \multicolumn{2}{c}{\textbf{Ours}} \\
\cmidrule(lr){2-3} \cmidrule(lr){4-5}
 & \textbf{Qustion Language} & \textbf{Score} & \textbf{Question Language} & \textbf{Score} \\
\midrule
\multirow{3}{*}{Culture} 
& English & 0.9976 & English & \textbf{0.7336} \\
& Japanese & 0.9707 & Japanese & \textbf{0.6007} \\
& Chinese & 0.9653 & Chinese & \textbf{0.6368} \\
\midrule
\multirow{4}{*}{Translation} 
& Japanese $\rightarrow$ English & 0.9559 & Japanese $\rightarrow$ English & \textbf{0.6180} \\
& Chinese $\rightarrow$ English & 0.9273 & Chinese $\rightarrow$ English & \textbf{0.7807} \\
& English $\rightarrow$ Japanese & 0.9940 & English $\rightarrow$ Japanese & \textbf{0.6925} \\
& English $\rightarrow$ Chinese & 0.9950 & English $\rightarrow$ Chinese & \textbf{0.6887} \\
\bottomrule
\end{tabular}
\label{tab:self-bleu}
\end{table}

\subsection{Mathmetical Approach for LogitLens}
\label{app:mathmetical-approach-for-logitlens}
We provide a detailed explanation of LogitLens~\cite{nostalgebraist2020logitlens} using mathematical notation.
Here, we define each variable used in Equations (1)--(3) for the LogitLens analysis.
In Equation (1),

$$
\mathbf{h}_{\ell+1} = \mathbf{h}_\ell + \mathcal{F}_\ell(\mathbf{h}_\ell)
$$

$\mathbf{h}_\ell \in \mathbb{R}^d$ is the hidden state at layer $\ell$, and $\mathcal{F}_\ell(\cdot)$ is the residual streams.
This equation describes the standard residual update in Transformer models.

In Equation (2),
\[
\mathcal{M}_{>\ell}(\mathbf{h}_\ell) =
\mathbf{W}_U^\top \,\,\mathrm{LayerNorm} \left[
\mathbf{h}_\ell +
\sum_{\ell' = \ell}^{L} \mathcal{F}_{\ell'}(\mathbf{h}_{\ell'})
\right]
\]

$\mathcal{M}_{>\ell}(h_\ell)$ denotes the final logits if forward computation continues from layer $\ell$.
$W_U \in \mathbb{R}^{d \times |V|}$ is the unembedding matrix, and $|V|$ is the vocabulary size. The sum represents all residual updates after layer $\ell$.

In Equation (3),

\[
\mathrm{LogitLens}(\mathbf{h}_\ell) =
\mathbf{W}_U^\top \, \mathrm{LayerNorm}(\mathbf{h}_\ell)
\]

The LogitLens removes all future residuals and directly maps $h_\ell$ to vocabulary logits. 
It shows what the model ``knows'' at layer $\ell$, independently of later layers.

\subsection{Detailed Models Settings}
\label{app:detailed-model-settings}
Details of the models used in the experiments are listed in Table~\ref{tab:detailed-model-names}. To ensure reproducibility, the seed was set to 42, and top\_p was fixed at 0.0. The implementation utilized the Transformers library~\cite{wolf-etal-2020-transformers} and bitsandbytes~\cite{dettmers2022gpt3}.
Note that all code and datasets will be available upon acceptance.

\begin{table}[h!]
    \caption{Detailed model names, parameters, and max input tokens.}
    \centering
    \small
    \begin{tabular}{lccl}
        \toprule
        \textbf{Model}  & \textbf{Params} & \textbf{Max Tokens} &\textbf{HuggingFace Name / OpenAI Name}  \\
        \midrule
        \texttt{Gemma3} & \texttt{1B} & 32,768 & \texttt{google/gemma-3-1b-it}  \\
        \texttt{Qwen2.5} & \texttt{1.5B} & 32,768 & \texttt{Qwen/Qwen2.5-1.5B-Instruct}  \\
        \texttt{LLM-jp-3} & \texttt{1.8B} & 4,096 & \texttt{llm-jp/llm-jp-3-1.8b-instruct}  \\
        \texttt{Qwen2.5 (3B)} & \texttt{3B} & 32,768 & \texttt{Qwen/Qwen2.5-3B-Instruct}  \\
        \texttt{LLM-jp-3 (7.2B)} & \texttt{7.2B} & 4,096 & \texttt{llm-jp/llm-jp-3-7.2b-instruct}  \\
        \texttt{GPT-4o} & --  & 128k & \texttt{gpt-4o-2024-11-20} \\
        \bottomrule
    \end{tabular}
    \label{tab:detailed-model-names}
\end{table}

\subsection{Details of Input Token Length}
\label{app:details-of-input-token-length}
As mentioned in Section~\ref{adversarial-prompts-settings}, our study aimed to fill the maximum input length of the model with adversarial prompts.
The number of tokens used for this purpose is listed in Table~\ref{tab:each-token-count-table} below.
In addition to these token counts, the input also includes 4-shot examples and the question text.

\begin{table}[h!]
    \caption{Adversarial Prompt Token Counts for Different Ratios}
    \small
    \centering
    \begin{tabular}{cc ccccc c}
    \toprule
    \multirow{2.25}{*}{\textbf{Model}} & \multirow{2.25}{*}{\textbf{Lang}} & \multicolumn{5}{c}{\textbf{Ratio}} & \multirow{2.25}{*}{\textbf{Max Token Length}} \\
    \cmidrule{3-7}
     & & \textbf{0.2} & \textbf{0.4} & \textbf{0.6} & \textbf{0.8} & \textbf{1.0} &  \\
    \midrule
    \multirow{3}{*}{LLM-jp-3} & Ja & 845 & 1,664 & 2,483 & 3,301 & 3,925 & \multirow{3}{*}{4,096} \\
                           & En & 842 & 1,661 & 2,479 & 3,298 & 3,921 & \\
                           & Zh & 848 & 1,667 & 2,485 & 3,304 & 3,928 &  \\
                           \cmidrule(lr){2-8}
    \multirow{3}{*}{Qwen2.5} & Ja & 6,584 & 13,138 & 19,691 & 26,245 & 32,631 & \multirow{3}{*}{32,768} \\
                           & En & 6,570 & 13,119 & 19,670 & 26,220 & 32,602 &  \\
                           & Zh & 65,71 & 13,125 & 19,678 & 26,232 & 32,618 &  \\
                           \cmidrule(lr){2-8}
    \multirow{3}{*}{Gemma3}  & Ja & 6,578 & 13,130 & 19,681 & 26,233 & 32,617 & \multirow{3}{*}{32,768} \\
                           & En & 6,571 & 13,121 & 19,671 & 26,221 & 32,604 & \\
                           & Zh & 6,571 & 13,122 & 19,672 & 26,222 & 32,605 & \\
    \bottomrule
    \end{tabular}
    \label{tab:each-token-count-table}
\end{table}

\subsection {Inference Settings}
\label{app:inference-settings}
In our study, as far as inference which needs to use GPUs, all experiments were conducted on a single NVIDIA RTX A6000 and NVIDIA A100 80GB PCIe.

\subsection{AI Assistant Tools}
\label{app:ai-assistant-tools}
We used ChatGPT~\footnote{\url{https://chatgpt.com/}} and DeepL~\footnote{\url{https://www.deepl.com/ja/translator}} to translate sentences to English and accelerate our research.

\subsection{Detailed Prompts}
\label{app:detailed-prompts}
Below are examples of the prompts used in the experiments.
As previously mentioned, the experiments were conducted with 4-shots, following prior studies.
The placeholder \{variable\} will be replaced with actual questions.
Due to their large size, the overall adversarial prompts will be published as URLs upon acceptance.

\subsubsection{Question Generation}
\begin{tcolorbox}[title=Prompt for geo-culture and translation task, boxrule=1pt, colback=white, coltitle=black, colframe=gray, colbacktitle=gray!10]
\small
\texttt{\textcolor{red}{\{Adversarial prompts\}} (if needed)} \\ \\

\texttt{「晴れ」を中国語に翻訳してください。, answer: 晴天} \\
\texttt{What type of government does the USA have?, answer: democracy} \\
\texttt{「春」を英語に翻訳してください。, answer: spring} \\
\texttt{What is the capital city of Japan?, answer: Tokyo} \\

\texttt{\textcolor{blue}{\{Question\}}, answer: }
\end{tcolorbox}

\begin{tcolorbox}[title=Prompt for creating question with related to geo-culture task by gpt-4o, boxrule=1pt, colback=white, coltitle=black, colframe=gray, colbacktitle=gray!10]
\small
  \texttt{Create a question related to the geography of \textcolor{blue}{\{country\}}.\\}
  \texttt{Please generate with related to \textcolor{blue}{\{category\}}.\\}
  \texttt{The question MUST consist of a single sentence.\\}

  \texttt{\# ** Important **\\}
  \texttt{- The answer format must follow the format below "\# Example".\\}
  \texttt{- Only 1 sentence is needed for the response.\\}
  \texttt{- Create questions where the answer will be only one word.\\}
  \texttt{- Please generate the question end with ", answer: " even though language is Japanese or Chinese.\\}

  \texttt{\# Example:\\}
  \texttt{What is the capital of \textcolor{blue}{\{country\}}?, answer:\\}
  \texttt{What is the highest mountain in \textcolor{blue}{\{country\}}?, answer:\\}
  \texttt{What is the official language of \textcolor{blue}{\{country\}}?, answer:\\}
  \texttt{What is the currency of \textcolor{blue}{\{country\}}?, answer:\\}
\end{tcolorbox}

\begin{tcolorbox}[title=Prompt for creating question with related to translation task by gpt-4o, boxrule=1pt, colback=white, coltitle=black, colframe=gray, colbacktitle=gray!10]
\small
  \texttt{Create a question related to the translation task from \textcolor{blue}{\{src\_lang\}} to \textcolor{blue}{\{trg\_lang\}}.\\}
  \texttt{Please generate the question in \textcolor{blue}{\{src\_lang\}} with related to \textcolor{blue}{\{category\}}.\\}

  \texttt{For example, you can create a question like "Please translate "Water" into Chinese., answer: ".\\}
  \texttt{This example is that the question is in English, whereas the answer will be Chinese.\\}
  \texttt{The question MUST ONLY consist of a single sentence.\\}

  \texttt{\# ** Important **\\}
  \texttt{- INPUT IS \textcolor{blue}{\{src\_lang\}}.\\}
  \texttt{- Please say "Please translate XXX into \textcolor{blue}{\{trg\_lang\}}, answer:"\\}
  \texttt{- The generation format must follow the format below "\# Example".\\}
  \texttt{- Only 1 sentence is needed for the response.\\}
  \texttt{- Create questions where the answer will be only one word.\\}
  \texttt{- Do NOT include an ANSWER in the generation.\\}
  \texttt{- Please generate the sentence end with ", answer: " EVEN IF Chinese or Japanese.\\}

  \texttt{\# Example:\\}
  \texttt{"東京"を英語に翻訳してください。, answer:\\}
  \texttt{Please translate "Beijing" into Chinese., answer:\\}
  \texttt{Please translate "Washington" into Japanese., answer:\\}
  \texttt{Please translate "Sun" into Japanese., answer:\\}
\end{tcolorbox}

\subsubsection{Adversarial Prompts}
\begin{tcolorbox}[title=Adversarial prompts in Japanese, boxrule=1pt, colback=white, coltitle=black, colframe=gray, colbacktitle=gray!10]
\small
\texttt{あなたは下記に示す日本人のように振る舞ってください。また日本人の思考で考えてください。}\\\\

\texttt{日本の文化や背景、特徴、歴史について説明します。}\\
\texttt{日本は、東アジアに位置する島国で、太平洋に面し、主に本州、北海道、九州、四国の4つの主要な島と、それに付随する約6,800以上の小さな島々から構成されています。国土の多くが山地で、自然災害、特に地震や台風が多い地域でもあります。これらの自然環境は、日本人の生活様式や精神性、信仰に大きな影響を与えてきました。}\\
\texttt{まず、日本の歴史について触れます。日本の歴史は数千年にわたり、時代ごとに独自の文化が形成されてきました。}\\
...... (Much more long sentences.) \\\\

\texttt{上記の内容を基に日本人のように振る舞ってください。}
\end{tcolorbox}

\begin{tcolorbox}[title=Adversarial prompts in Chinese, boxrule=1pt, colback=white, coltitle=black, colframe=gray, colbacktitle=gray!10]
\small
\Chinese{请像中国人一样表现。} \\\\

\Chinese{中国，全称为中华人民共和国，是位于東亞的一个幅员辽阔、历史悠久、文化多元的国家。它拥有世界上最多的人口（约14亿），以及广阔的领土，横跨东西约5000公里，南北跨度也极大，地理、气候和文化的多样性使得中国成为世界上最为复杂和丰富的文明体之一。以下将从中国的文化、背景、特点与历史等方面进行尽可能详细的描述。}\\
...... (Much more long sentences.) \\\\

\Chinese{在提及上述内容时，请表现得像个中国人}
\end{tcolorbox}

\begin{tcolorbox}[title=Adversarial prompts in English, boxrule=1pt, colback=white, coltitle=black, colframe=gray, colbacktitle=gray!10]
\small
\texttt{Please behave like a US person readin below sentences. Please think in English native speaker.} \\

\texttt{The United States of America (USA) is a vast and diverse country located in North America, bordered by Canada to the north, Mexico to the south, the Atlantic Ocean to the east, and the Pacific Ocean to the west. The U.S. is a federal republic composed of 50 states, a federal district (Washington, D.C.), and several territories. Its enormous size, cultural plurality, complex history, and global influence make it one of the most important and influential countries in the modern world.} \\
\texttt{Let us now explore its culture, background, characteristics, and history in as much detail as possible.} \\
...... (Much more long sentences.) \\\\

\texttt{Please behave like a US person referring to the above content.}
\end{tcolorbox}

\subsection{Examples of Questions}
\label{app:examples-of-questions}
Below are examples of questions generated using the method described in Section~\ref{dataset-creation}.
Note that (answer) is excluded during input.

\begin{tcolorbox}[title=Examples of Questions in translation task, boxrule=1pt, colback=white, coltitle=black, colframe=gray, colbacktitle=gray!10]
\small
\texttt{「晴れ」を中国語に翻訳してください。, answer: (晴天)} \\
\texttt{「一」を英語に翻訳してください。, answer: (one)} \\
\texttt{\Chinese{请将“星期一”翻译成日语}, answer: (月曜日)} \\
\texttt{\Chinese{"请将“交通”翻译成英语}, answer: (transportation)} \\
\texttt{"キャベツ"を中国語に翻訳してください。, answer: (卷心菜)}
\end{tcolorbox}

\begin{tcolorbox}[title=Examples of Questions in geo-culture task, boxrule=1pt, colback=white, coltitle=black, colframe=gray, colbacktitle=gray!10]
\small
\texttt{Who is the current Prime Minister of Japan?, answer: (Kishida)} \\
\texttt{現在の日本の首相は誰ですか？, answer: (岸田)} \\
\texttt{\Chinese{日本目前的首相是谁？， answer: (岸田)}} \\ \\

\texttt{What is the capital city of China?, answer: (Beijing)} \\
\texttt{中国の首都はどこですか?, answer: (北京)} \\ 
\texttt{\Chinese{中国的首都是什么？, answer: (北京)}} \\ \\

\texttt{What is considered the national dish of the United States?, answer: (Hamburger)} \\

\texttt{アメリカ合衆国の国民食と見なされるのは何ですか?, answer: (ハンバーガー)} \\
\texttt{\Chinese{么被认为是美国的国菜?, answer: (汉堡)}} \\
\end{tcolorbox}

\subsection{Other Results}
\label{app:other-results}
Below are additional experimental results.
As these results indicate, it is not possible to determine whether the inclusion of adversarial prompts directly affects consistency based on this table.
However, it is evident that robustness and confidence tend to decline.

\begin{figure*}[h!]
  \begin{minipage}[b]{0.5\linewidth}
    \centering
    \includegraphics[width=\linewidth]{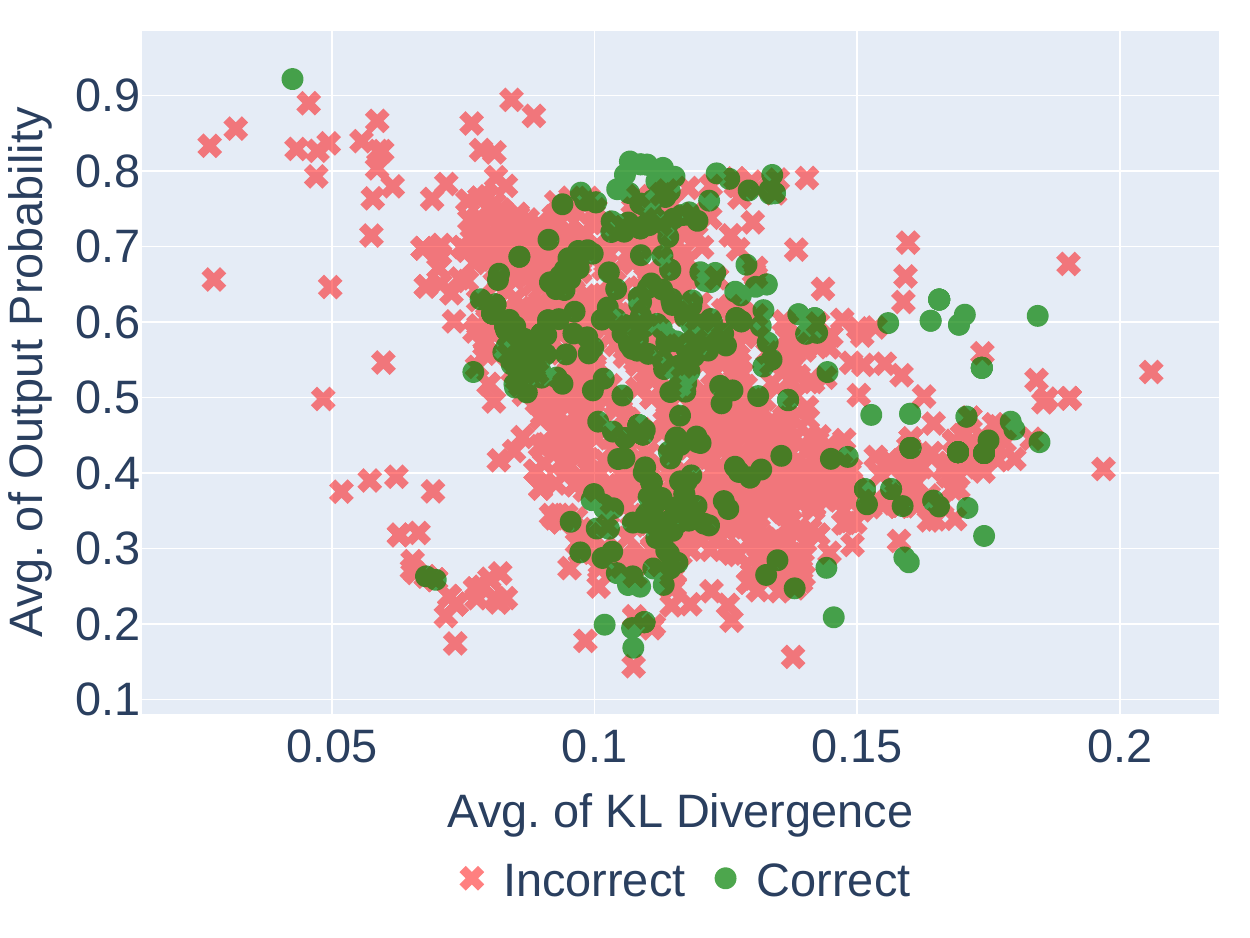}
    \caption*{(a) Question in Japanese w/o adversarial}
  \end{minipage}
  \begin{minipage}[b]{0.5\linewidth}
    \centering
    \includegraphics[width=\linewidth]{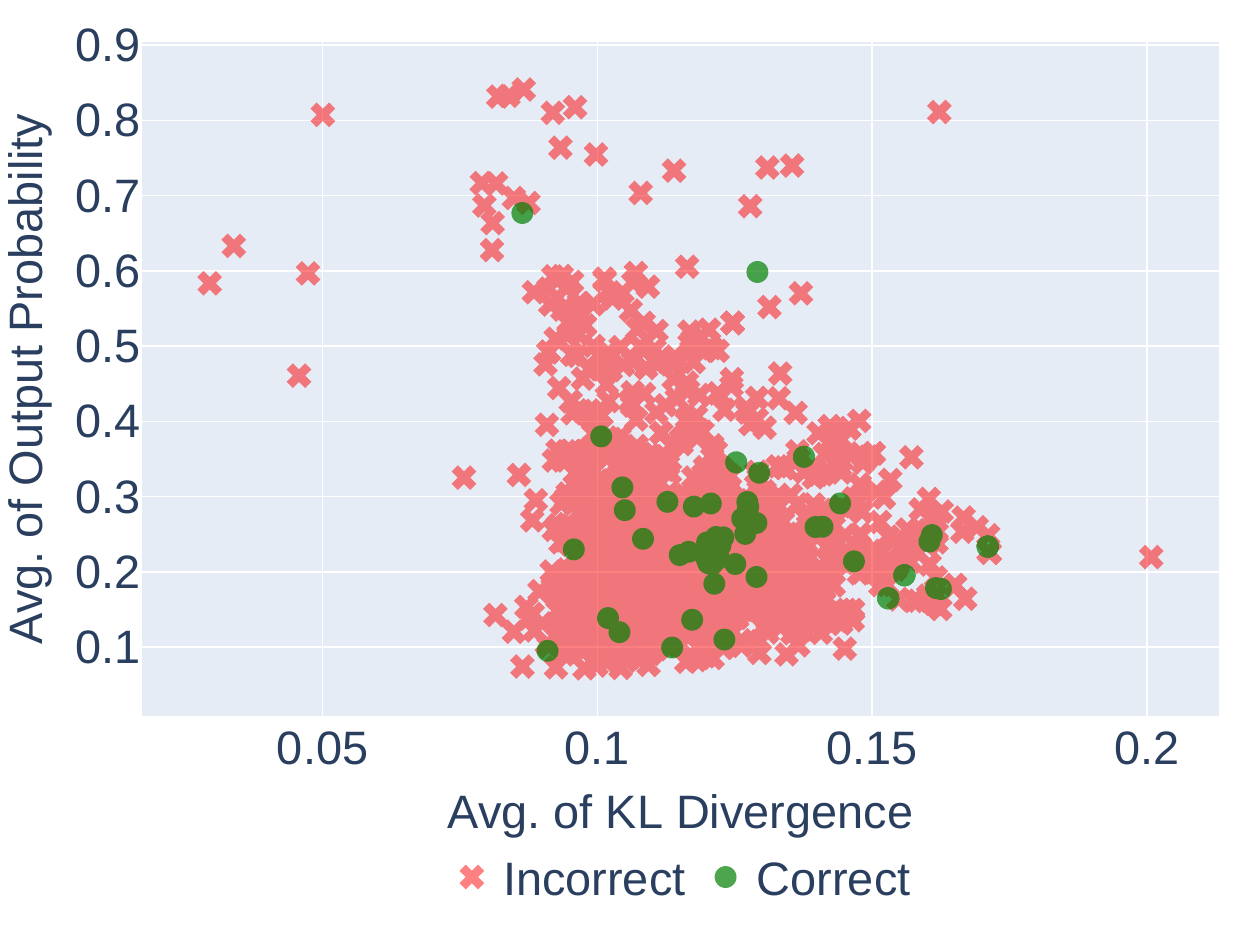}
    \caption*{(b) Question and adversarial in Japanese}
  \end{minipage}
\caption{
Results of the LLM-jp-3 model on the geo-culture task. The left panel shows results without noise, while the right panel shows results with 100\% noise.
Out of 2,000 samples, green points indicate correct predictions, while red points indicate incorrect ones.
}
  \label{data-plot-geo-other-results1}
\end{figure*}

\begin{figure*}[h!]
  \begin{minipage}[b]{0.5\linewidth}
    \centering
    \includegraphics[width=\linewidth]{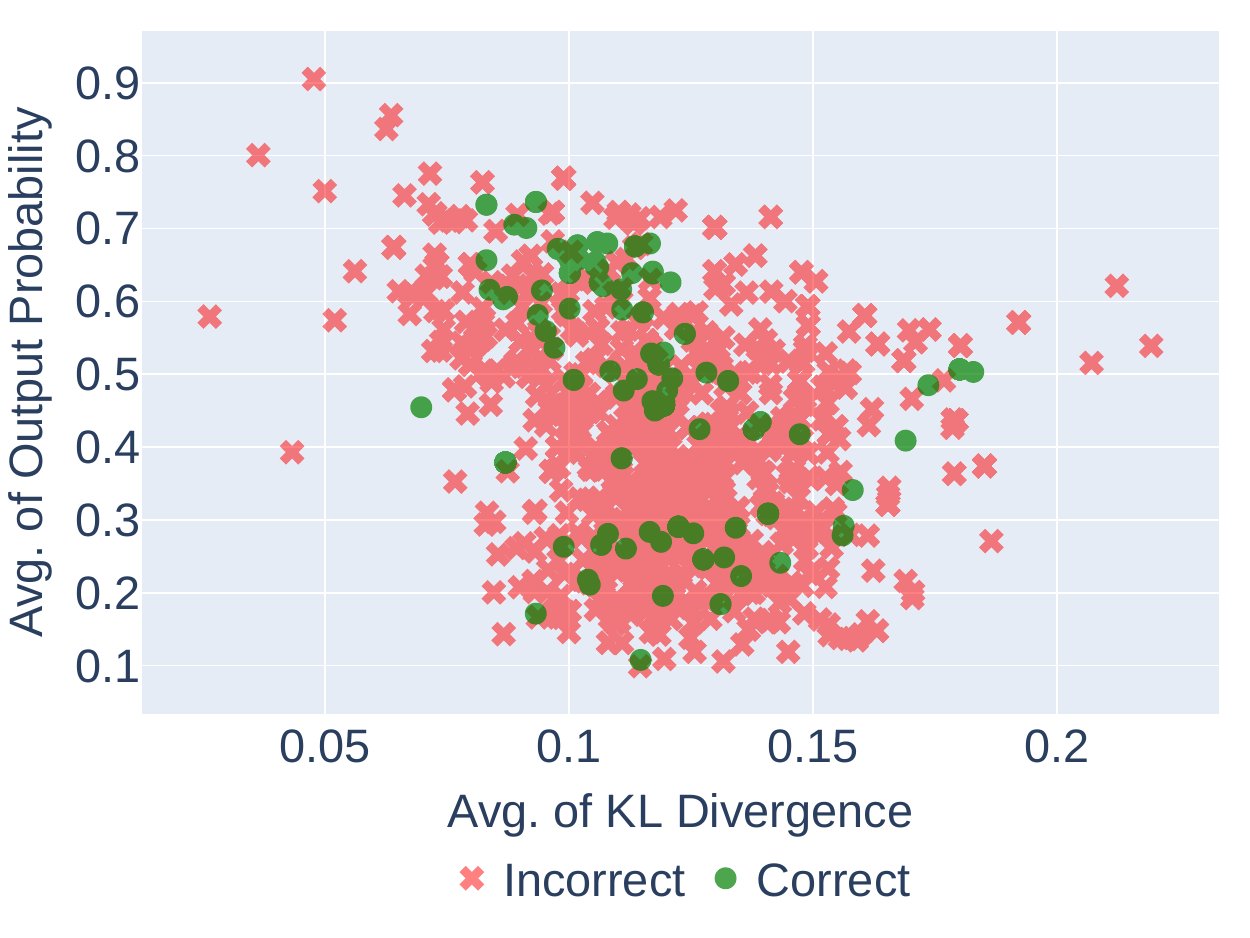}
    \caption*{(a) Question in Japanese w/o adversarial}
  \end{minipage}
  \begin{minipage}[b]{0.5\linewidth}
    \centering
    \includegraphics[width=\linewidth]{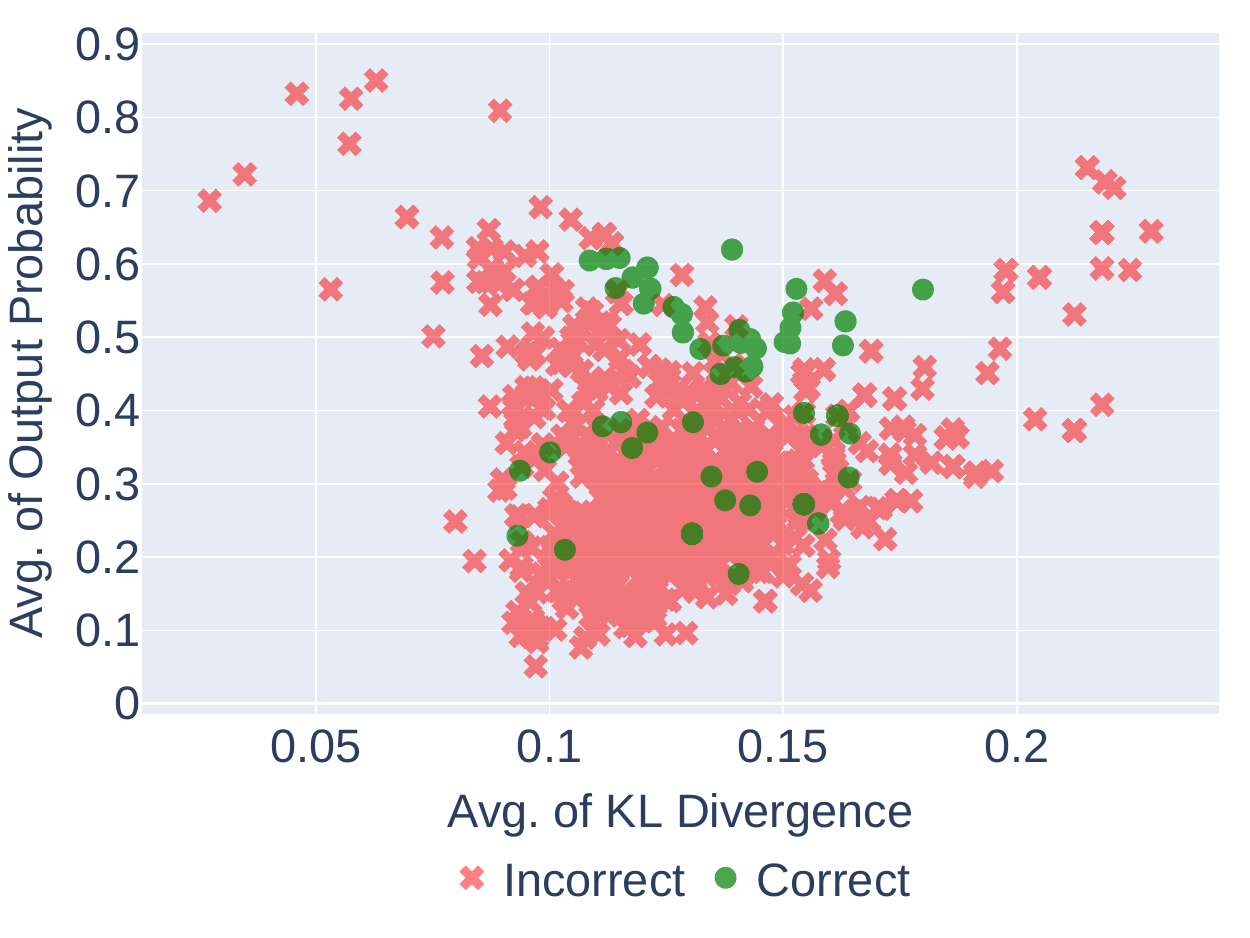}
    \caption*{(b) Question and adversarial in Japanese}
  \end{minipage}
\caption{
Results of Qwen2.5 on the geo-culture task.
The figure is interpreted in the same way as Figure~\ref{data-plot-geo-other-results1}.
 }
  \label{data-plot-geo-other-results2}
\end{figure*}

\begin{table}[t]
\caption{Correlation between LLC Score and robustness in translation tasks. The value $r$ represents the correlation coefficient for each noise ratio.}
    \centering
    \resizebox{\textwidth}{!}{
    \begin{tabular}{ccccc ccccc ccccc}
    \toprule
\multirow{2.25}{*}{\textbf{Model}} & \multirow{2.25}{*}{\textbf{Source}} & \multirow{2.25}{*}{\textbf{Target}} & \multirow{2.25}{*}{\textbf{Adversarial}} & \multicolumn{5}{c}{\textbf{LLC Score ($\downarrow$)}} & \multicolumn{5}{c}{\textbf{Robustness ($\uparrow$)}} & \multirow{2.25}{*}{\(r\)} \\
\cmidrule(lr){5-9}\cmidrule(lr){10-14}
 & & & & \textbf{0.2} & \textbf{0.4} & \textbf{0.6} & \textbf{0.8} & \textbf{1.0} & \textbf{0.2} & \textbf{0.4} & \textbf{0.6} & \textbf{0.8} & \textbf{1.0} \\
\midrule
\multirow{12}{*}{LLM-jp-3} & \multirow{3}{*}{Ja} & \multirow{3}{*}{En} & Ja & $\text{En}_{0.07}$ & $\text{En}_{0.07}$ & $\text{Ja}_{0.13}$ & $\text{Ja}_{0.07}$ & $\text{Ja}_{0.08}$ & 0.78 & 0.77 & 0.73 & 0.71 & 0.74 & -0.37 \\
 & & & En & $\text{En}_{0.07}$ & $\text{Zh}_{0.12}$ & $\text{Ja}_{0.09}$ & $\text{Ja}_{0.09}$ & $\text{Ja}_{0.46}$ & 0.81 & 0.76 & 0.68 & 0.69 & 0.61 & -0.74 \\
 & & & Zh & $\text{En}_{0.08}$ & $\text{En}_{0.07}$ & $\text{Zh}_{0.09}$ & $\text{En}_{0.07}$ & $\text{En}_{0.10}$ & 0.77 & 0.76 & 0.74 & 0.66 & 0.55 & -0.56 \\
\cmidrule(lr){3-15}
 & \multirow{6}{*}{En} & \multirow{3}{*}{Ja} & Ja & $\text{Zh}_{0.12}$ & $\text{Zh}_{0.12}$ & $\text{Zh}_{0.11}$ & $\text{En}_{0.14}$ & $\text{En}_{0.16}$ & 0.73 & 0.71 & 0.69 & 0.59 & 0.37 & -0.96 \\
 & & & En & $\text{Ja}_{0.12}$ & $\text{Ja}_{0.15}$ & $\text{Ja}_{0.14}$ & $\text{Ja}_{0.12}$ & $\text{En}_{0.25}$ & 0.70 & 0.67 & 0.67 & 0.57 & 0.62 & -0.21 \\
 & & & Zh & $\text{En}_{0.10}$ & $\text{Ja}_{0.19}$ & $\text{En}_{0.11}$ & $\text{Zh}_{0.14}$ & $\text{En}_{0.20}$ & 0.62 & 0.66 & 0.57 & 0.56 & 0.55 & 0.02 \\
 \cmidrule(lr){3-15}
 & & \multirow{3}{*}{Zh} & Ja & $\text{En}_{0.06}$ & $\text{En}_{0.06}$ & $\text{En}_{0.08}$ & $\text{En}_{0.10}$ & $\text{Ja}_{0.16}$ & 0.33 & 0.33 & 0.32 & 0.28 & 0.19 & -1.00 \\
 & & & En & $\text{En}_{0.09}$ & $\text{En}_{0.06}$ & $\text{Zh}_{0.12}$ & $\text{En}_{0.15}$ & $\text{Ja}_{0.10}$ & 0.37 & 0.35 & 0.31 & 0.29 & 0.27 & -0.59 \\
 & & & Zh & $\text{En}_{0.06}$ & $\text{En}_{0.06}$ & $\text{En}_{0.15}$ & $\text{En}_{0.14}$ & $\text{En}_{0.15}$ & 0.39 & 0.38 & 0.30 & 0.29 & 0.17 & -0.85 \\
 \cmidrule(lr){3-15}
 & \multirow{3}{*}{Zh} & \multirow{3}{*}{En} & Ja & $\text{Ja}_{0.08}$ & $\text{En}_{0.08}$ & $\text{En}_{0.11}$ & $\text{Zh}_{0.12}$ & $\text{En}_{0.10}$ & 0.81 & 0.79 & 0.74 & 0.72 & 0.60 & -0.42 \\
 & & & En & $\text{En}_{0.08}$ & $\text{Zh}_{0.12}$ & $\text{En}_{0.08}$ & $\text{En}_{0.07}$ & $\text{En}_{0.09}$ & 0.80 & 0.79 & 0.65 & 0.42 & 0.49 & 0.48 \\
 & & & Zh & $\text{En}_{0.09}$ & $\text{En}_{0.09}$ & $\text{Ja}_{0.09}$ & $\text{Ja}_{0.12}$ & $\text{En}_{0.08}$ & 0.71 & 0.71 & 0.72 & 0.70 & 0.57 & 0.29 \\
 \cmidrule(lr){2-15}
\multirow{12}{*}{Qwen2.5} & \multirow{3}{*}{Ja} & \multirow{3}{*}{En} & Ja & $\text{En}_{0.00}$ & $\text{En}_{0.00}$ & $\text{En}_{0.00}$ & $\text{En}_{0.00}$ & $\text{En}_{0.00}$ & 0.84 & 0.85 & 0.85 & 0.84 & 0.84 & N.A. \\
 & & & En & $\text{En}_{0.00}$ & $\text{En}_{0.00}$ & $\text{En}_{0.00}$ & $\text{En}_{0.05}$ & $\text{En}_{0.05}$ & 0.75 & 0.76 & 0.74 & 0.65 & 0.67 & -0.99 \\
 & & & Zh & $\text{En}_{0.00}$ & $\text{En}_{0.00}$ & $\text{En}_{0.00}$ & $\text{En}_{0.00}$ & $\text{En}_{0.00}$ & 0.79 & 0.78 & 0.73 & 0.69 & 0.69 & N.A. \\
 \cmidrule(lr){3-15}
 & \multirow{6}{*}{En} & \multirow{3}{*}{Ja} & Ja & $\text{En}_{0.00}$ & $\text{En}_{0.00}$ & $\text{En}_{0.00}$ & $\text{En}_{0.00}$ & $\text{En}_{0.00}$ & 0.00 & 0.00 & 0.00 & 0.00 & 0.00 & N.A. \\
 & & & En & $\text{En}_{0.00}$ & $\text{En}_{0.00}$ & $\text{En}_{0.00}$ & $\text{En}_{0.00}$ & $\text{En}_{0.00}$ & 0.00 & 0.00 & 0.00 & 0.00 & 0.00 & N.A. \\
 & & & Zh & $\text{En}_{0.00}$ & $\text{En}_{0.00}$ & $\text{En}_{0.00}$ & $\text{En}_{0.00}$ & $\text{En}_{0.00}$ & 0.00 & 0.00 & 0.00 & 0.00 & 0.00 & N.A. \\
 \cmidrule(lr){3-15}
 & & \multirow{3}{*}{Zh} & Ja & $\text{En}_{0.00}$ & $\text{En}_{0.00}$ & $\text{En}_{0.00}$ & $\text{En}_{0.00}$ & $\text{En}_{0.00}$ & 0.02 & 0.01 & 0.00 & 0.00 & 0.00 & N.A. \\
 & & & En & $\text{En}_{0.00}$ & $\text{En}_{0.00}$ & $\text{En}_{0.00}$ & $\text{En}_{0.00}$ & $\text{En}_{0.00}$ & 0.01 & 0.01 & 0.02 & 0.00 & 0.00 & N.A. \\
 & & & Zh & $\text{En}_{0.00}$ & $\text{En}_{0.00}$ & $\text{En}_{0.00}$ & $\text{En}_{0.00}$ & $\text{En}_{0.00}$ & 0.01 & 0.01 & 0.00 & 0.00 & 0.00 & N.A. \\
\cmidrule(lr){3-15}
 &  \multirow{3}{*}{Zh} & \multirow{3}{*}{En} & Ja & $\text{En}_{0.00}$ & $\text{En}_{0.08}$ & $\text{En}_{0.06}$ & $\text{En}_{0.09}$ & $\text{En}_{0.09}$ & 0.46 & 0.50 & 0.33 & 0.55 & 0.54 & 0.38 \\
 & & & En & $\text{En}_{0.00}$ & $\text{En}_{0.08}$ & $\text{En}_{0.09}$ & $\text{En}_{0.08}$ & $\text{En}_{0.00}$ & 0.68 & 0.61 & 0.57 & 0.65 & 0.75 & -0.86 \\
 & & & Zh & $\text{En}_{0.00}$ & $\text{En}_{0.00}$ & $\text{En}_{0.00}$ & $\text{En}_{0.06}$ & $\text{En}_{0.07}$ & 0.85 & 0.83 & 0.78 & 0.73 & 0.74 & -0.88 \\
\cmidrule(lr){2-15}
\multirow{12}{*}{Gemma3} & \multirow{3}{*}{Ja} & \multirow{3}{*}{En} & Ja & $\text{En}_{0.02}$ & $\text{En}_{0.02}$ & $\text{En}_{0.02}$ & $\text{En}_{0.02}$ & $\text{En}_{0.02}$ & 0.78 & 0.77 & 0.76 & 0.30 & 0.59 & 0.76 \\
 & & & En & $\text{En}_{0.06}$ & $\text{En}_{0.06}$ & $\text{En}_{0.03}$ & $\text{En}_{0.06}$ & $\text{En}_{0.03}$ & 0.80 & 0.82 & 0.78 & 0.73 & 0.68 & 0.49 \\
 & & & Zh & $\text{En}_{0.03}$ & $\text{En}_{0.08}$ & $\text{En}_{0.07}$ & $\text{En}_{0.24}$ & $\text{En}_{0.10}$ & 0.80 & 0.79 & 0.69 & 0.24 & 0.47 & -0.92 \\
 \cmidrule(lr){3-15}
 & \multirow{6}{*}{En} & \multirow{3}{*}{Ja} & Ja & $\text{En}_{0.05}$ & $\text{En}_{0.02}$ & $\text{En}_{0.05}$ & $\text{En}_{0.01}$ & $\text{En}_{0.00}$ & 0.27 & 0.14 & 0.14 & 0.08 & 0.07 & 0.83 \\
 & & & En & $\text{En}_{0.05}$ & $\text{En}_{0.05}$ & $\text{En}_{0.05}$ & $\text{En}_{0.00}$ & $\text{En}_{0.00}$ & 0.30 & 0.18 & 0.20 & 0.11 & 0.14 & 0.81 \\
 & & & Zh & $\text{En}_{0.06}$ & $\text{En}_{0.05}$ & $\text{En}_{0.05}$ & $\text{En}_{0.05}$ & $\text{En}_{0.05}$ & 0.32 & 0.28 & 0.17 & 0.17 & 0.15 & 0.77 \\
 \cmidrule(lr){3-15}
 & & \multirow{3}{*}{Zh} & Ja & $\text{En}_{0.05}$ & $\text{En}_{0.04}$ & $\text{Zh}_{0.02}$ & $\text{En}_{0.04}$ & $\text{En}_{0.04}$ & 0.22 & 0.16 & 0.15 & 0.05 & 0.04 & 0.30 \\
 & & & En & $\text{En}_{0.08}$ & $\text{En}_{0.05}$ & $\text{En}_{0.04}$ & $\text{En}_{0.01}$ & $\text{En}_{0.01}$ & 0.23 & 0.15 & 0.14 & 0.02 & 0.02 & 0.98 \\
 & & & Zh & $\text{En}_{0.10}$ & $\text{En}_{0.06}$ & $\text{En}_{0.08}$ & $\text{En}_{0.05}$ & $\text{En}_{0.05}$ & 0.24 & 0.24 & 0.13 & 0.13 & 0.06 & 0.59 \\
 \cmidrule(lr){3-15}
 & \multirow{3}{*}{Zh} & \multirow{3}{*}{En} & Ja & $\text{En}_{0.00}$ & $\text{En}_{0.00}$ & $\text{En}_{0.00}$ & $\text{En}_{0.01}$ & $\text{En}_{0.01}$ & 0.84 & 0.81 & 0.79 & 0.47 & 0.55 & -0.99 \\
 & & & En & $\text{En}_{0.03}$ & $\text{En}_{0.03}$ & $\text{En}_{0.00}$ & $\text{En}_{0.02}$ & $\text{En}_{0.02}$ & 0.82 & 0.85 & 0.73 & 0.71 & 0.55 & 0.33 \\
 & & & Zh & $\text{En}_{0.00}$ & $\text{En}_{0.00}$ & $\text{En}_{0.01}$ & $\text{En}_{0.03}$ & $\text{En}_{0.00}$ & 0.76 & 0.62 & 0.48 & 0.37 & 0.14 & -0.23 \\
     \bottomrule
    \end{tabular}
    } 
    \label{tab:result-table-translation}
\end{table}

\clearpage
\subsection{Results by Larger LLMs}
\label{app:results-by-larger-llms}
In our study, we used relatively small LLMs for evaluation, as described in Appendix~\ref{tab:detailed-model-names}.
This choice was made to ensure that inference with maximum noise relative to token length remains feasible on a single GPU.
Larger models present significant challenges in terms of computational resources, making such evaluations difficult.
Nevertheless, we conducted experiments with larger models, including Qwen (3B) and LLM-jp-3 (7.2B), by reducing the noise ratio.
As shown in Table~\ref{tab:result-larger-table-geo} and Table~\ref{tab:result-larger-table-translation}, these models exhibited similar performance trends to their smaller counterparts.
Future work aims to extend our work to larger models with more extensive noise injection using multiple GPUs.

\begin{figure*}[h!]
  \begin{minipage}[b]{0.5\linewidth}
    \centering
    \includegraphics[width=\linewidth]{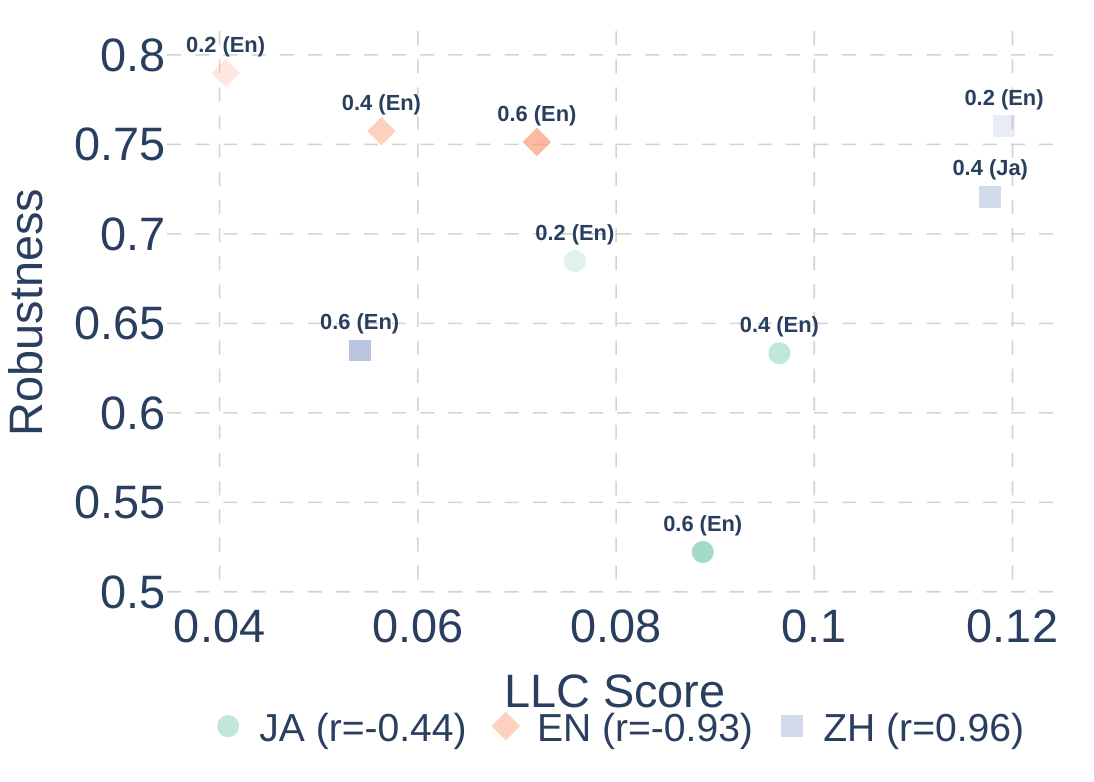}
    \caption*{(a) Src: En, Trg: Ja}
  \end{minipage}
  \begin{minipage}[b]{0.5\linewidth}
    \centering
    \includegraphics[width=\linewidth]{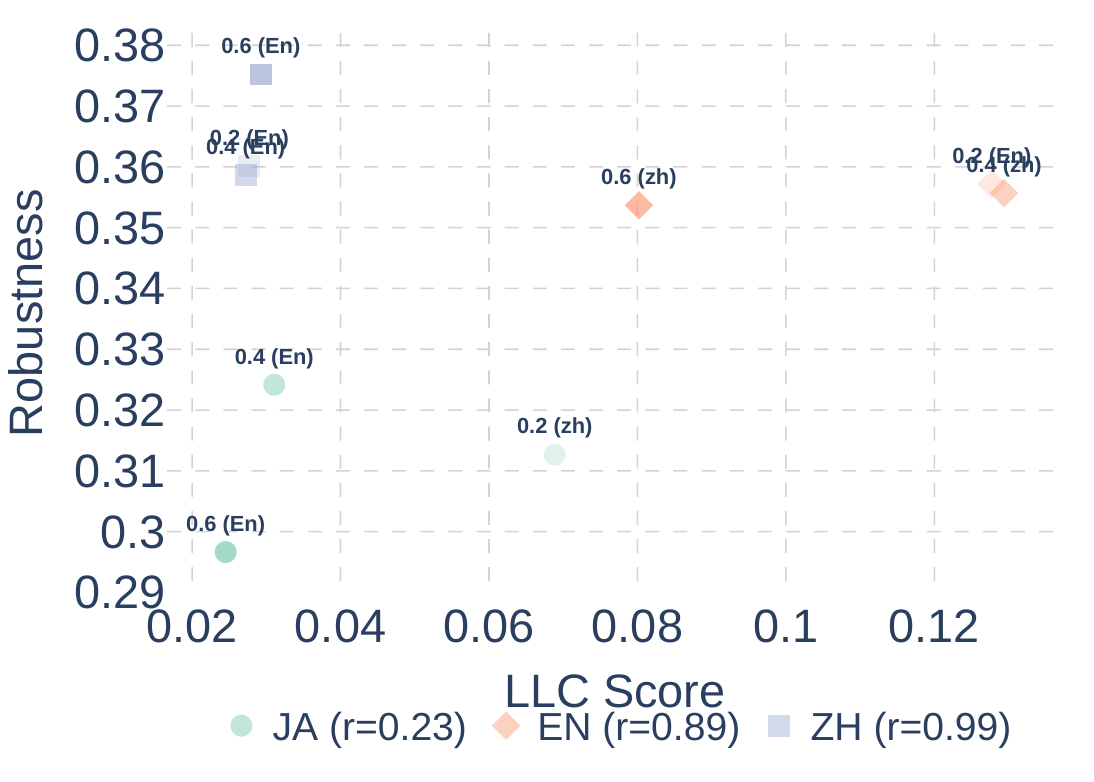}
    \caption*{(b) Src: En, Trg: Zh}
  \end{minipage}
\caption{
Correlation between language consistency and robustness in translation tasks using LLM-jp-3 (7.2B).
The color of each point represents the adversarial language, indicating the ratio and the dominant language in each setting.
}
\label{app:larger-model-results-translation-llm-jp}
\end{figure*}

\begin{table}[h!]
\caption{
Correlation between LLC Score and robustness in geo-culture tasks.
The value $r$ represents the correlation coefficient for each noise ratio.
}
    \small
    \centering
    \resizebox{\textwidth}{!}{
    \begin{tabular}{cccc ccc ccc}
    \toprule
\multirow{2.25}{*}{\textbf{Model}} & \multirow{2.25}{*}{\textbf{Question}} & \multirow{2.25}{*}{\textbf{Adversarial}} & \multicolumn{3}{c}{\textbf{LLC Score ($\downarrow$)}} & \multicolumn{3}{c}{\textbf{Robustness ($\uparrow$)}} & \multirow{2.25}{*}{\textbf{\(r\)}} \\
\cmidrule(lr){4-6}\cmidrule(lr){7-9}
 & & &\textbf{0.2} & \textbf{0.4} & \textbf{0.6} & \textbf{0.2} & \textbf{0.4} & \textbf{0.6} \\
\midrule
\multirow{8.5}{*}{\shortstack{LLM-jp-3 \\ (7.2B)}} & \multirow{2}{*}{Ja} & Ja & $\text{Ja}_{0.13}$ & $\text{Ja}_{0.12}$ & $\text{En}_{0.07}$ & 0.26 & 0.26 & 0.29 & -0.97 \\
 & & En & $\text{Ja}_{0.12}$ & $\text{Ja}_{0.12}$ & $\text{Ja}_{0.09}$ & 0.28 & 0.28 & 0.27 & 1.00 \\
  \cmidrule(lr){3-10}
 & \multirow{3}{*}{En} & Ja & $\text{En}_{0.12}$ & $\text{En}_{0.10}$ & $\text{En}_{0.10}$ & 0.11 & 0.16 & 0.17 & -1.00 \\
 & & En & $\text{En}_{0.08}$ & $\text{En}_{0.08}$ & $\text{En}_{0.08}$ & 0.23 & 0.23 & 0.23 & 0.75 \\
 & & Zh & $\text{En}_{0.11}$ & $\text{En}_{0.10}$ & $\text{En}_{0.10}$ s& 0.11 & 0.19 & 0.19 & -0.90 \\
  \cmidrule(lr){3-10}
 & \multirow{2}{*}{Zh} & En & $\text{Ja}_{0.11}$ & $\text{Ja}_{0.09}$ & $\text{Ja}_{0.07}$ & 0.08 & 0.08 & 0.03 & 0.89 \\
 & & Zh & $\text{En}_{0.12}$ & $\text{En}_{0.12}$ & $\text{Ja}_{0.09}$ & 0.07 & 0.05 & 0.07 & -0.56 \\
 \cmidrule(lr){2-10}
\multirow{8.5}{*}{\shortstack{Qwen2.5 \\ (3B)}} & \multirow{2}{*}{Ja} & Ja & $\text{En}_{0.12}$ & $\text{En}_{0.12}$ & $\text{En}_{0.12}$ & 0.00 & 0.00 & 0.00 & -0.05 \\
 &  & En & $\text{En}_{0.09}$ & $\text{En}_{0.10}$ & $\text{En}_{0.10}$ & 0.00 & 0.00 & 0.00 & N.A. \\
 \cmidrule(lr){3-10}
 & \multirow{3}{*}{En} & Ja & $\text{En}_{0.09}$ & $\text{En}_{0.09}$ & $\text{En}_{0.09}$ & 0.32 & 0.29 & 0.34 & -1.00 \\
 & & En & $\text{En}_{0.08}$ & $\text{En}_{0.09}$ & $\text{En}_{0.09}$ & 0.23 & 0.17 & 0.20 & -0.95 \\
 & & Zh & $\text{En}_{0.09}$ & $\text{En}_{0.09}$ & $\text{En}_{0.09}$ & 0.28 & 0.27 & 0.29 & -0.95 \\
\cmidrule(lr){3-10}
 & \multirow{2}{*}{Zh} & En & $\text{En}_{0.10}$ & $\text{Zh}_{0.04}$ & $\text{En}_{0.10}$ & 0.01 & 0.01 & 0.01 & 0.62 \\
 & & Zh & $\text{Zh}_{0.04}$ & $\text{Zh}_{0.04}$ & $\text{Zh}_{0.04}$ & 0.00 & 0.01 & 0.01 & 0.94 \\
    \bottomrule
    \end{tabular}
    } 
    \label{tab:result-larger-table-geo}
\end{table}

\begin{table}[ht]
\caption{Correlation between LLC Score and robustness in translation tasks. The value $r$ represents the correlation coefficient for each noise ratio.}
    \small
    \centering
    \resizebox{\textwidth}{!}{
    \begin{tabular}{ccccc ccc ccc}
    \toprule
\multirow{2.25}{*}{\textbf{Model}} & \multirow{2.25}{*}{\textbf{Source}} & \multirow{2.25}{*}{\textbf{Target}} & \multirow{2.25}{*}{\textbf{Adversarial}} & \multicolumn{3}{c}{\textbf{LLC Score ($\downarrow$)}} & \multicolumn{3}{c}{\textbf{Robustness ($\uparrow$)}} & \multirow{2.25}{*}{\textbf{\(r\)}} \\
\cmidrule(lr){5-7}\cmidrule(lr){8-10}
 & & & & \textbf{0.2} & \textbf{0.4} & \textbf{0.6} & \textbf{0.2} & \textbf{0.4} & \textbf{0.6}  \\
\midrule
\multirow{13.5}{*}{\shortstack{LLM-jp-3 \\ (7.2B)}} & \multirow{3}{*}{Ja} & \multirow{3}{*}{En} & Ja & $\text{En}_{0.07}$ & $\text{En}_{0.07}$ & $\text{En}_{0.07}$ & 0.78 & 0.75 & 0.73 & -0.27 \\
 & & & en & $\text{Ja}_{0.11}$ & $\text{En}_{0.12}$ & $\text{En}_{0.13}$ & 0.77 & 0.74 & 0.74 & -0.66 \\
 & & & Zh & $\text{Ja}_{0.10}$ & $\text{Ja}_{0.12}$ & $\text{Ja}_{0.12}$ & 0.72 & 0.69 & 0.72 & -0.48 \\
  \cmidrule(lr){3-11}
 & \multirow{3}{*}{En} & \multirow{3}{*}{Ja} & ja & $\text{En}_{0.08}$ & $\text{En}_{0.10}$ & $\text{En}_{0.09}$ & 0.68 & 0.63 & 0.52 & -0.44 \\
 & & & En & $\text{En}_{0.04}$ & $\text{En}_{0.06}$ & $\text{En}_{0.07}$ & 0.79 & 0.76 & 0.75 & -0.93 \\
 & & & Zh & $\text{En}_{0.12}$ & $\text{Ja}_{0.12}$ & $\text{En}_{0.05}$ & 0.76 & 0.72 & 0.63 & 0.96 \\
  \cmidrule(lr){3-11}
 & \multirow{3}{*}{En} & \multirow{3}{*}{Zh} & Ja & $\text{Zh}_{0.07}$ & $\text{En}_{0.03}$ & $\text{En}_{0.02}$ & 0.31 & 0.32 & 0.30 & 0.23 \\
 & & & En & $\text{En}_{0.13}$ & $\text{Zh}_{0.13}$ & $\text{Zh}_{0.08}$ & 0.36 & 0.36 & 0.35 & 0.89 \\
 & & & zh & $\text{En}_{0.03}$ & $\text{En}_{0.03}$ & $\text{En}_{0.03}$ & 0.36 & 0.36 & 0.38 & 0.99 \\
   \cmidrule(lr){3-11}
 & \multirow{3}{*}{Zh} & \multirow{3}{*}{En} & Ja & $\text{Zh}_{0.10}$ & $\text{En}_{0.13}$ & $\text{En}_{0.13}$ & 0.72 & 0.73 & 0.71 & 0.04 \\
 & & & en & $\text{Ja}_{0.08}$ & $\text{En}_{0.10}$ & $\text{En}_{0.13}$ & 0.78 & 0.69 & 0.71 & -0.70 \\
 & & & Zh & $\text{Zh}_{0.08}$ & $\text{En}_{0.14}$ & $\text{En}_{0.12}$ & 0.75 & 0.73 & 0.75 & -0.73 \\
 \cmidrule(lr){2-11}
\multirow{13.5}{*}{\shortstack{Qwen2.5 \\ (3B)}} & \multirow{3}{*}{Ja} & \multirow{3}{*}{En} & Ja & $\text{En}_{0.05}$ & $\text{Zh}_{0.05}$ & $\text{Zh}_{0.05}$ & 0.87 & 0.86 & 0.87 & -0.98 \\
 & & & en & $\text{Zh}_{0.04}$ & $\text{Zh}_{0.05}$ & $\text{Zh}_{0.05}$ & 0.85 & 0.83 & 0.83 & -0.95 \\
 & & & Zh & $\text{En}_{0.05}$ & $\text{En}_{0.06}$ & $\text{En}_{0.05}$ & 0.85 & 0.88 & 0.86 & 0.98 \\
 \cmidrule(lr){3-11}
 & \multirow{3}{*}{En} & \multirow{3}{*}{Ja} & Ja & $\text{zh}_{0.05}$ & $\text{En}_{0.05}$ & $\text{zh}_{0.05}$ & 0.02 & 0.02 & 0.04 & -0.95 \\
 & & & En & $\text{zh}_{0.04}$ & $\text{zh}_{0.04}$ & $\text{zh}_{0.04}$ & 0.00 & 0.00 & 0.00 & -0.98 \\
 & & & Zh & $\text{En}_{0.06}$ & $\text{En}_{0.05}$ & $\text{En}_{0.04}$ & 0.02 & 0.00 & 0.00 & 0.84 \\
  \cmidrule(lr){3-11}
 & \multirow{3}{*}{En} & \multirow{3}{*}{Zh} & Ja & $\text{zh}_{0.05}$ & $\text{zh}_{0.05}$ & $\text{zh}_{0.05}$ & 0.02 & 0.02 & 0.02 & -0.80 \\
 & & & En & $\text{zh}_{0.04}$ & $\text{En}_{0.05}$ & $\text{zh}_{0.04}$ & 0.02 & 0.02 & 0.02 & -0.35 \\
 & & & zh & $\text{En}_{0.05}$ & $\text{Zh}_{0.05}$ & $\text{Zh}_{0.05}$ & 0.02 & 0.02 & 0.02 & 1.00 \\
 \cmidrule(lr){3-11}
 & \multirow{3}{*}{Zh} & \multirow{3}{*}{En} & Ja & $\text{En}_{0.06}$ & $\text{Zh}_{0.05}$ & $\text{Zh}_{0.05}$ & 0.87 & 0.86 & 0.86 & 1.00 \\
 & & & en & $\text{Zh}_{0.05}$ & $\text{Zh}_{0.05}$ & $\text{Zh}_{0.05}$ & 0.85 & 0.86 & 0.88 & 0.83 \\
 & & & Zh & $\text{Zh}_{0.06}$ & $\text{En}_{0.05}$ & $\text{Zh}_{0.06}$ & 0.89 & 0.88 & 0.89 & 0.61 \\
     \bottomrule
    \end{tabular}
    } 
    \label{tab:result-larger-table-translation}
\end{table}


\end{document}